\documentclass[10pt]{article} 
\usepackage{graphicx} 

\usepackage{microtype}
\usepackage{graphicx}
\usepackage{subfigure}

\usepackage{booktabs} 
\usepackage[table]{xcolor}

\usepackage{hyperref}
\usepackage{caption}
\usepackage{textgreek}

\usepackage[margin=1in]{geometry}

\usepackage{amsmath}
\usepackage{amssymb}
\usepackage{mathtools}
\usepackage{amsthm}
\usepackage{comment}
\usepackage[capitalize,noabbrev]{cleveref}

\usepackage{algorithm}
\usepackage{algcompatible}

\usepackage{url}

\theoremstyle{plain}

\theoremstyle{remark}

\usepackage[margin=1in]{geometry}

\usepackage[textsize=tiny]{todonotes}
\usepackage{natbib}
\usepackage{blindtext}
\usepackage[utf8]{inputenc} 
\usepackage[T1]{fontenc}    
\usepackage{url}            
\usepackage{booktabs}       
\usepackage{multirow}       
\usepackage{amsfonts}       
\usepackage{nicefrac}       
\usepackage{microtype}      
\long\def\comment#1{}

\title{
    \vspace{-2cm}
    \includegraphics[width=100pt]{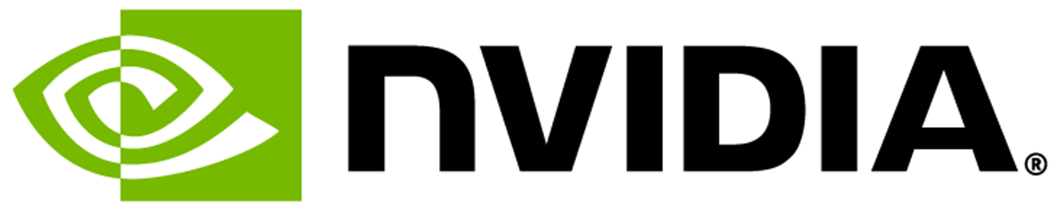}\\[1em]
    Nemotron-Labs-3-Puzzle-75B-A9B:\\Compressing Hybrid MoE LLMs
}
\date{}

\begin{document}
\maketitle
\author{Akhiad Bercovich, Talor Abramovich, Daniel Afrimi, Shay Aharon, Nir Ailon, Vladimir Anisimov, Omer Ullman Argov, Maor Ashkenazi, Tomer Asida, Nave Assaf, Tomer Bar Natan,  Alexander Bukharin, Grzegorz Chlebus, Marcin Chochowski, Eric Chung, Mohammad Dabbah, Carlo del Mundo, Ewa Dobrowolska, Ido Galil, Yaniv Galron, Amnon Geifman, Yonatan Geifman, Izik Golan, Alex Gronskiy, Tomasz Grzegorzek, Netanel Haber, Lior Kadoch, Grzegorz Karch, Tomer Keren, Abhinav Khattar, Amir Klein, Tugrul Konuk, Roi Koren, Daniel Korzekwa, Shaun Kotek, Konstantinos Krommydas, Itay Levy , Ofri Masad, Yoav Miron, Pavlo Molchanov, Shahar Mor, Zach Moshe, Saurav Muralidharan, Najeeb Nabwani, Besmira Nushi, Mostofa Patwary, Omri Puny, Johannes Rausch, Tomer Ronen, Sepehr Sameni, Itamar Schen, Elad Segal, Daniel Serebrenik, Ido Shahaf, Soumye Singhal, Daniil Sorokin, Sharath Turuvekere Sreenivas, Marta Stepniewska-Dziubinska, Ali Taghibakhshi, Nima Tajbakhsh, Oren Tropp, Dor Tzur, Anna Warno, Yi-Fu Wu, Michal Zawalski, Jiaqi Zeng, Yian Zhang, Ran Zilberstein, Amit Zuker, Ran El-Yaniv}

\begin{abstract}
We present Nemotron-Labs-3-Puzzle-75B-A9B, a compressed variant of Nemotron-3-Super optimized for interactive deployment. We designed the model to maximize server throughput under high user throughput constraints.

In interactive serving workloads on a single 8×B200 node, Puzzle-75B-A9B achieves approximately 2× higher server throughput than Nemotron-3-Super at matched user throughput constraints.

In ultra-long-context deployment on a single H100 GPU, the compressed model increases 1M-token concurrency from 1 request to 8 requests.

Puzzle-75B-A9B is constructed using a multi-stage pipeline that combines the Iterative Puzzle compression framework with knowledge distillation, reinforcement learning, quantization,
and Multi-Token Prediction head.
The compression process jointly optimizes heterogeneous MoE pruning, active parameter budget, and Mamba pruning to improve inference efficiency while preserving model quality.

We evaluate Puzzle-75B-A9B on a broad suite of reasoning, coding, multilingual, long-context, and agentic benchmarks. Despite substantial compression, the model retains strong downstream accuracy relative to the parent model across a wide range of tasks.  These results demonstrate that large hybrid MoE models can be substantially optimized for deployment efficiency while maintaining strong downstream capability. Our model is publicly available on Hugging Face \footnote{The models are available at:
\href{https://huggingface.co/nvidia/NVIDIA-Nemotron-Labs-3-Puzzle-75B-A9B-BF16}{NVIDIA-Nemotron-Labs-3-Puzzle-75B-A9B-BF16},
\href{https://huggingface.co/nvidia/NVIDIA-Nemotron-Labs-3-Puzzle-75B-A9B-NVFP4}{NVIDIA-Nemotron-Labs-3-Puzzle-75B-A9B-NVFP4}, and
\href{https://huggingface.co/nvidia/NVIDIA-Nemotron-Labs-3-Puzzle-75B-A9B-FP8}{NVIDIA-Nemotron-Labs-3-Puzzle-75B-A9B-FP8}.}.

\end{abstract}

\begin{figure}[t]
    \centering
    \includegraphics[width=0.7\textwidth]{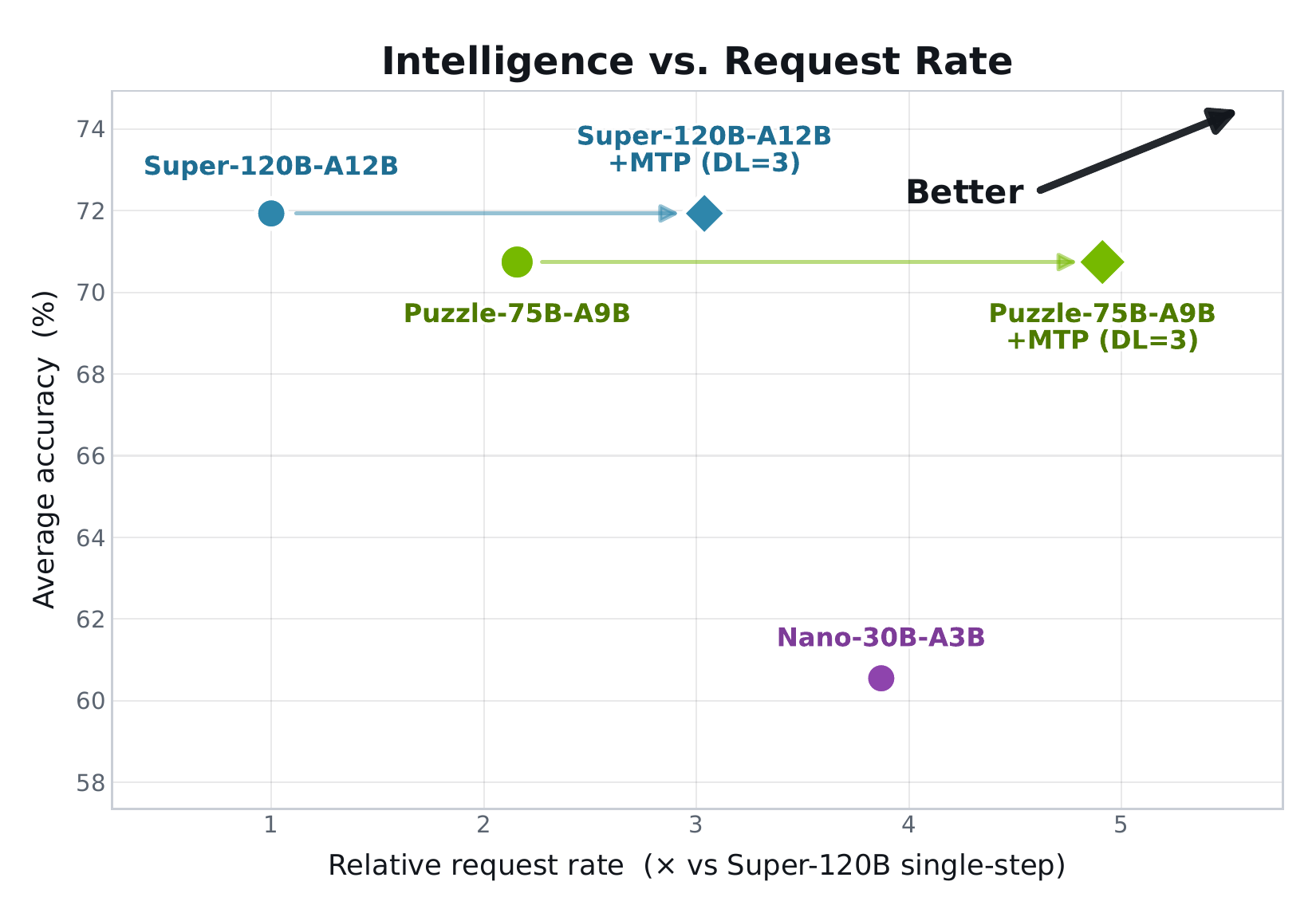}
    \caption{Accuracy--efficiency tradeoff on a single 8$\times$B200 node with all models served at matched
NVFP4 quantization. The x-axis is the relative number of completed
requests, normalized to Nemotron-3-Super single-step
decoding $=$ 1. It is computed from Pareto-optimal total throughput
at user throughput $\mathrm{UT}{=}100$ tok/s and then
divided by each model's average generation verbosity. The y-axis is the unweighted
suite-average accuracy across the benchmarks in
Table~\ref{tab:super_main_results}. }
    \label{fig:accuracy_speed_frontier_b200}
    
\end{figure}
\section{Introduction}
\label{sec:intro}

Recent large language models (LLMs) have achieved substantial gains in reasoning, coding, long-context understanding, and agentic behavior by scaling both parameter count and architectural complexity. However, these improvements come at a significant inference cost,  high memory consumption, elevated serving latency, and reduced deployment flexibility.  These limitations become particularly acute in production inference settings, where systems must simultaneously satisfy strict throughput, latency, and context-length requirements. Consequently, improving inference efficiency without substantially degrading model quality has emerged as a central challenge for deploying next-generation LLMs at scale.

A common approach to this problem is post-training compression, where a pretrained model is transformed into a smaller and more efficient model. Existing compression techniques, including pruning, quantization, and knowledge distillation, often involve a difficult tradeoff between efficiency and downstream capability preservation. This challenge is especially pronounced for hybrid MoE architectures, where performance depends not only on total parameter count, but also on active parameters, KV-cache size, Mamba-state size, and the ratio of input/output sequence lengths.
Moreover, different deployment scenarios have different bottlenecks: requests with long inputs are typically compute-bound, requests with long outputs are typically IO-bound (unless the batch size is very large), and ultra-long-context requests are often restricted by device memory due to their large KV-cache footprint. 

In this work, we present \emph{Nemotron-Labs-3-Puzzle-75B-A9B}, a compressed and deployment-optimized variant of Nemotron-3-Super~\cite{nemotron_3_super}, designed to substantially improve inference efficiency while preserving the parent model’s capabilities across a wide range of tasks, including reasoning, coding, and long-context. Our approach combines architecture-aware compression with targeted post-training recovery techniques. At the core of our method is \emph{Iterative Puzzle}, a sequential extension of the Puzzle framework~\cite{puzzle}  that alternates between hardware-aware structural compression and short recovery phases based on knowledge distillation (KD). Rather than compressing the model in a single step, Iterative Puzzle progressively constructs a sequence of intermediate architectures, enabling the compression process to adapt to changing internal representations and accounting for inter-layer dependencies or interactions.

The resulting compression pipeline jointly optimizes multiple architectural dimensions of the hybrid model. In particular, we perform heterogeneous channel pruning of MoE routed experts, heterogeneous active experts reduction, and prune the SSM state size of Mamba layers. Unlike uniform scaling approaches, the proposed method allocates capacity non-uniformly across network depth according to measured layer importance and deployment constraints. This enables the compressed architecture to preserve capacity in the most sensitivity-critical regions of the network while aggressively reducing redundancy elsewhere. To further recover capabilities degraded by compression, we apply large-scale KD over both medium- and long-context training regimes, followed by reinforcement learning (RL) post-training focused on software-engineering and agentic reasoning tasks. We additionally combine the compressed architecture with deployment-oriented quantization and speculative decoding techniques, including FP8/NVFP4 mixed-precision inference and multi-token prediction (MTP).

We target two representative deployment settings that reflect common production workloads: (1) interactive serving on a single NVIDIA B200 node with 8 GPUs, where per-user throughput is the primary user-experience constraint, and (2) ultra-long-context agentic workloads on a single NVIDIA H100 GPU, where memory footprint becomes the dominant bottleneck due to KV-cache scaling at million-token contexts. Accordingly, we optimize Puzzle-75B-A9B for both a prefill-heavy 50K/2K (input/output) regime characteristic of long-context analysis and retrieval-augmented generation (RAG), and a decode-heavy 8K/64K regime representative of reasoning-oriented generation. For interactive serving, our target is a 2× server throughput improvement over the parent model at an operating point of at least 100 tokens per second for each user (100 TPS user throughput).

For ultra-long-context deployment, we target sustainable concurrency of 8 simultaneous 1M-token requests on a single H100 GPU, compared to the parent model’s effective limit of 1 request under the combined KV-cache, Mamba-state memory and parameter memory budget. 

The final Nemotron-Labs-3-Puzzle-75B-A9B model compresses the Nemotron-3-Super from 120.7B total/12.8B active parameters to 75.3B total/9.3B active parameters, achieving substantial efficiency gains while preserving strong performance across reasoning, coding, multilingual, long-context, and agentic benchmarks.
Moreover, compared to Nemotron-3-Nano~\cite{nemotron3-nano}, which has 30B total/3.5B active parameters, Figure~\ref{fig:accuracy_speed_frontier_b200} shows Puzzle-75B-A9B is highly efficient.  Specifically, when accounting for Nano’s higher generation verbosity, a server running Puzzle-75B-A9B+MTP at 100 TPS user throughput completes as many user requests per minute as a server running Nano without MTP, while achieving near-Super accuracy. We note that additional throughput gains may also be achievable for Nano through the use of a dedicated MTP module.

\section{Methodology: Compression and Recovery Techniques}

Our methodology consists of a two-stage pipeline:  structural compression followed by performance recovery. In the first stage, we compress the original Nemotron-3-Super model into a smaller, deployment-efficient architecture by introducing an iterative variant of the Puzzle framework. Specifically, the proposed Iterative Puzzle scheme alternates between progressive structural compression following the principles of \cite{puzzle} and short phases of knowledge distillation (KD). This iterative approach enables stable and effective exploration of the architecture under hardware-aware constraints, allowing systematic optimization for target inference environments.

In the second stage, we address the performance degradation typically introduced by aggressive compression. To this end, we apply a combination of knowledge distillation (KD) and reinforcement learning (RL) fine-tuning techniques to recover (and in some cases improve) the model’s accuracy on a wide range of tasks. Knowledge distillation transfers the predictive behavior of the original large model to the compressed student model, while RL-based optimization further aligns the model with downstream objectives and model behavior. Together, these “recovery” techniques enable the compressed model to maintain high accuracy while benefiting from significantly reduced computational and memory requirements. 

To further improve inference speed we use quantization and speculative decoding. We produce deployment-oriented NVFP4 and FP8 checkpoints using post-training quantization (PTQ) recipes tailored to Blackwell- and Hopper-class GPUs, respectively.  We also leverage the native Multi-Token Prediction (MTP) capabilities inherited from Nemotron-3-Super. We show that the original shared-head MTP transfers effectively to the compressed architecture, and further improve its robustness through continued training designed to mitigate the mismatch between teacher-forced MTP training and autoregressive speculative decoding at inference time. Together, these optimizations substantially improve deployment efficiency while preserving strong downstream capability.

\subsection{Puzzle Phase}

Puzzle \citep{puzzle} is a decomposed \emph{neural architecture search} (NAS) framework for LLMs.
Given a trained ``\emph{parent model}'', Puzzle searches for a derivative architecture that satisfies deployment efficiency constraints (e.g., memory footprint, latency, and throughput) while preserving the parent’s accuracy.
It does so by (i) defining a discrete search space of alternative layer implementations (or ``puzzle pieces''), (ii) estimating and assigning each alternative piece a quality score (based on its efficiency/accuracy profile), and (iii) solving a \emph{mixed-integer program} (MIP) to select one alternative piece per layer under the target constraints. We use the Puzzletron~\cite{puzzletron} implementation to run these steps.

\textbf{Search space.} Nemotron-3-Super is composed of Mamba, Attention and Mixture-of-Experts (MoE) layers. We focus our search on different MoE pruning strategies for the MoE layers, using Puzzle to build a heterogeneous architecture that allocates different capacities to each MoE layer according to its estimated impact on the model's accuracy, while satisfying our overall runtime targets. We prune the Mamba layers uniformly across the model due to inference framework limitations, and leave the Attention layers untouched since Nemotron-3-Super is already very kv-cache efficient. Additional search-space components explored during development but not included in the final model are described in the appendix.

\subsubsection{Pruning Techniques}
We consider multiple pruning strategies for MoE and Mamba layers.

\textbf{Intermediate channel pruning:} We rank intermediate channels within each routed expert based on their contribution to the expert’s output, following the same criterion used for dense FFN pruning in Puzzle~\cite{puzzle}. Within each MoE layer, all routed experts are pruned to a uniform size to maintain compatibility with standard inference kernels, while allowing different layers to have varying intermediate dimensions. This form of pruning reduces both active and total parameter counts, improving memory efficiency and runtime performance across all deployment settings.

\textbf{Top-k reduction:} In each MoE layer, tokens are routed to $k$ selected experts for computing the next hidden state. We allow k to vary across layers, up to the original value of k=22 in the parent model. Reducing k decreases the number of active parameters per token, leading to improved efficiency in compute-bound scenarios such as prefill and large-batch decoding.

\textbf{Mamba SSM Pruning:} During the decode stage of text generation, cache IO is a large part of the Mamba layer runtime -- for Nemotron-3-Super at batch sizes 64 and above, cache IO takes the majority of the Mamba layer decode time.

In this work, we pruned the SSM state size of Nemotron-3-Super from 128 to 96 channels, resulting in a 1.2x-1.3x speedup of the SSM kernel during the decode stage for batch sizes between 8 and 512.

We chose which SSM channels to prune by estimating their contribution to the Mamba layer output -- detailed formulation and results in Appendix~\ref{appendix:ssm_pruning}.

\subsubsection{Iterative Puzzle}

The original Puzzle formulation makes the architecture search tractable by assuming that the quality impact of block replacements is approximately additive. In particular, each candidate block replacement is scored by replacing only the candidate block in the parent model and generating a metric-based score. The final architecture is selected by optimizing the sum of these local scores under a deployment constraint. While effective, this single-shot formulation does not explicitly model higher-order interactions between replacements. For example, pruning one MoE layer may change the distribution of representations seen by later layers, causing some blocks to become more important and others more redundant than their original replace-one scores suggest.

To mitigate this limitation, we introduce \emph{Iterative Puzzle}, a sequential compression procedure that alternates between moderate Puzzle-based pruning and short knowledge-distillation recovery phases. Instead of solving directly for the final target architecture, Iterative Puzzle constructs a sequence of progressively compressed models. After each Puzzle step, the resulting model is healed with short KD and then used as the starting point for the next round of scoring and optimization. This allows replacement scores to be recomputed in the context of the current compressed model and pruning choices already made in previous steps, rather than only in the context of the original parent model.

\textbf{Algorithm-} Let $M_0$ denote the original pretrained model and let $T$ denote the teacher model used for distillation. Iterative Puzzle constructs a sequence of models
\[
M_0 \rightarrow M_1 \rightarrow \cdots \rightarrow M_R,
\]
where each step applies a bounded amount of structural compression followed by a recovery stage.

At iteration $r$, we construct a Puzzle search space around the current model $M_{r-1}$ and evaluate candidate replacements in the context of this model. We then solve the Puzzle optimization problem under an incremental pruning budget, producing a compressed intermediate model $\widetilde{M}_r$.

Finally, we heal $\widetilde{M}_r$ using a short KD phase from the teacher $T$, and set the recovered model as the starting point for the next round:
\[
M_r = \mathrm{Heal}(\widetilde{M}_r, T).
\]
The process is repeated until the final deployment target is reached.

This iterative procedure preserves the decomposed structure that makes Puzzle scalable, while making the search more adaptive to the changing internal representations of the student model. In practice, we find that using moderate per-round compression budgets improves on single step puzzle in accuracy on downstream tasks for the same compression target. We describe the implementation details and final architecture in Sec.~\ref{sec:iter-puzz-implementation-details}, and present the ablation comparing Iterative Puzzle with single-step Puzzle in Sec.~\ref{sec:ablations}.

\subsection{Recovery Phase}
\textbf{Knowledge Distillation-} We apply knowledge distillation (KD) in both the Iterative Puzzle phase and the subsequent recovery phase. In both stages, training is performed on a mixed dataset comprising 30\% pretraining data and 70\% supervised fine-tuning (SFT) data from Nemotron-3-Nano \citep{nemotron3-nano}. During the Iterative Puzzle phase, each compression iteration is followed by a KD step in which the student model is trained to match the logits of the original Nemotron Super model. This distillation is performed until initial recovery is observed, with a sequence length of 32K, enabling efficient alignment at intermediate compression stages.

In the recovery phase, we extend distillation to longer contexts to further recover model performance. Training begins with a sequence length of 128K and is subsequently scaled to 512K tokens. As in the earlier phase, distillation is performed using logits from the Super model. For both sequence lengths, we use a training budget of up to 100B tokens and a global batch size of 16M tokens, implemented within the Megatron-LM framework~\citep{megatron-lm}.

\textbf{Reinforcement Learning-} Following the KD phase, we apply a reinforcement learning (RL) post-training stage aimed at recovering capabilities that are particularly sensitive to compression. Empirically, we observe that software engineering performance degrades most significantly, and thus we focus our RL efforts on restoring this capability.

To this end, we adopt Stage 2 of the Nemotron-3-Super RL pipeline \citep{nemotron_3_super} to target software engineering tasks. This stage is divided into two phases. Phase 2.1 (SWE 1) focuses on single-step tool-use comparison, generating 16 rollouts per prompt with a global batch size of 1024 and a maximum sequence length of 131,072. Phase 2.2 (SWE 2) transitions to full end-to-end sandbox RL utilizing isolated environments, where agents can execute up to 200 turns. In this phase, we generate 32 rollouts per prompt with a global batch size of 512, maintaining the maximum sequence length of 131,072. Both phases operate with a KL penalty of 0.

To further improve training stability and final model quality, we employ a checkpoint averaging strategy. Specifically, we repeat the SWE-RL training procedure  across multiple runs with a sweep over the learning rate ($1\times 10^{-7}$, $5\times 10^{-7}$, $1\times 10^{-6}$, and $5\times 10^{-6}$). We tried various averaging methods as proposed in \citet{model_soups}, but we found that regular weight averaging gave the best results. After convergence, we average the resulting model weights to produce the final model. This approach reduces variance across runs and leads to more stable and consistently high-performing policies.

\subsection{Implementation Details and Architecture}
\label{sec:iter-puzz-implementation-details}
We applied three compression-and-recovery stages. Each stage targeted a different axis.
In the first stage, we reduced the MoE weights to $75\%$ of the teacher capacity and pruned the Mamba SSM state size to $75\%$ of the teacher. The resulting model was healed for 24B tokens. In the second stage, we further reduced the MoE weights to $60\%$ of the teacher capacity and healed the model for 43.2B tokens. In the final stage, we constrained the activated routed-expert budget to $50\%$ of the teacher, allowing Puzzle to allocate this budget heterogeneously across layers. This final model was healed for 52.8B tokens.

The final model preserves the parent hybrid block layout, with 88 total blocks: 40 Mamba blocks, 40 MoE blocks, and 8 attention blocks. Table~\ref{tab:final-architecture} summarizes the dimensions that are changed by Iterative Puzzle relative to the parent model.

\begin{table}[t]
\centering
\small
\setlength{\tabcolsep}{5pt}
\begin{tabular}{lccc}
\toprule
Quantity & Super & Puzzle-75B-A9B & Puzzle-75B-A9B / Super \\
\midrule
Total parameters & 120.7B & 75.3B & 62.4\% \\
Active parameters & 12.8B & 9.3B & 73.1\% \\
\midrule
Mamba SSM state size & 128 & 96 & 75\% \\
MoE routed expert intermediate size & 2688 & 1280--2688 & Mean 59.9\% \\
Activated routed experts per token & 22 & 4--18 & Mean 50\% \\
Active routed experts capacity (relative) & 100\% & 8.7\%--62.3\% & Mean 30.9\% \\
\bottomrule
\end{tabular}
\caption{
Comparison between the parent model and the final compressed architecture. Parameter counts include embeddings, the LM head, and normalization parameters. The final architecture preserves the parent hybrid block layout, but prunes the Mamba SSM state size and assigns heterogeneous routed MoE capacity across layers. Quantities that are unchanged from the parent, such as the number of routed experts, shared expert size, and MoE latent size, are omitted.
}
\label{tab:final-architecture}
\end{table}

To visualize the layer-wise choices made by Puzzle, we plot the product of the two MoE dimensions that Puzzle was allowed to prune: the number of activated routed experts, $k_l$, and the routed expert intermediate size, $d_l$. This product is proportional to the active routed Expert parameters that each token sees in layer $l$. We normalize this quantity by the corresponding teacher value,
\[
\rho_l = \frac{k_l d_l}{k_T d_T},
\]
where $k_T$ and $d_T$ denote the teacher top-$k$ and routed experts intermediate size, respectively. Since the latent expert dimension is fixed across layers, $\rho_l$ captures the relative active routed-expert capacity allocated to each MoE layer, while excluding terms that are fixed across the compared MoE layers, such as the router, shared expert, and latent projections.

\begin{figure}[t]
    \centering
    \includegraphics[width=0.95\linewidth]{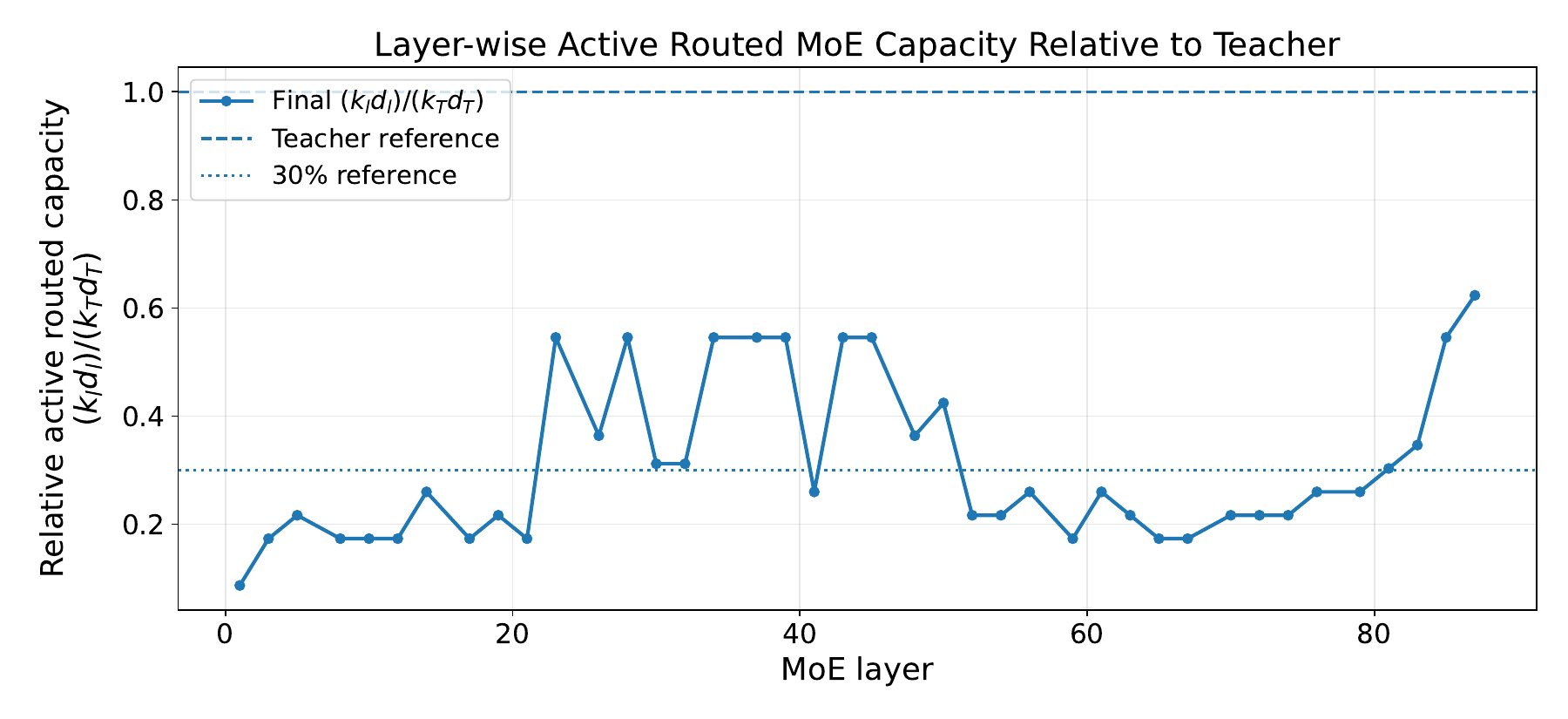}
    \caption{
    Layer-wise active routed Experts capacity relative to the teacher. For each MoE layer $l$, we plot
    $\rho_l = (k_l d_l)/(k_T d_T)$, where $k_l$ is the number of activated routed experts and $d_l$ is the routed expert intermediate size. These are the two MoE dimensions pruned by Puzzle, and their product is proportional to the active routed experts parameters seen by each token. The dashed horizontal line at $1.0$ denotes the teacher capacity, while the dotted line marks a $30\%$ reference, which represents the average $\rho$ across layers.
    }
    \label{fig:moe-relative-active-capacity}
\end{figure}

Figure~\ref{fig:moe-relative-active-capacity} shows that Puzzle selects a highly non-uniform allocation across depth. Rather than reducing every MoE layer to the same fraction of the teacher, the final architecture aggressively prunes many layers while preserving substantially more routed capacity in selected middle and late layers. This illustrates that the compressed model is not a uniformly scaled-down teacher, but a layer-wise architecture whose active MoE capacity is allocated according to measured layer sensitivity.

\subsection{Quantization}
\label{sec:quantization}
We apply post-training quantization (PTQ) to produce two quantized checkpoints: an FP8 W8A8 checkpoint targeting Hopper-class GPUs and an NVFP4 W4A4 checkpoint targeting Blackwell-class GPUs.

\paragraph{FP8 model.}
For the FP8 checkpoint, we follow the~\href{https://huggingface.co/nvidia/NVIDIA-Nemotron-3-Super-120B-A12B-FP8}{Nemotron-3-Super-120B-A12B-FP8} PTQ recipe. Calibration is
performed on a small subset of post-training SFT examples (256 samples, 4K tokens each),
and the primary compute-heavy operators are quantized to FP8. In particular, we quantize the
MoE GEMMs, including both routed and shared experts, and the Mamba linear layers. The
KV cache is quantized to FP8 in order to reduce memory overhead and to allow performing the first BMM of the scaled dot product attention operation in FP8 precision. The embedding, attention QKV
and attention output projections, MoE latent projections, Mamba 1D convolutions, and LM head, as well as the entire MTP, are kept in BF16, and the router is kept in FP32. We kept the Mamba SSM state in FP32 precision, since FP16 results in reduced accuracy unless stochastic rounding (SR) is employed, and we found that FP16+SR is slower than FP32 on Hopper (on Blackwell it is hardware supported).

\paragraph{NVFP4 model.} For the NVFP4 checkpoint, we used a similar approach. We calibrated again on a small subset of post-training SFT examples (256 samples, 65K tokens each).
The Nemotron-3-Super NVFP4 checkpoint ~\citep{nemotron_3_super}, used additional techniques, namely selecting per-block scales that minimize weight MSE and using Model-Optimizer AutoQuantize for selective quantization based on layer sensitivity.
We, instead, decided on the quantization level per layer type according to the majority trends seen in~\href{https://huggingface.co/nvidia/NVIDIA-Nemotron-3-Super-120B-A12B-NVFP4}{Nemotron-3-Super-120B-A12B-NVFP4} and used max calibration. Decisions are detailed in Table~\ref{tab:superturbo-quant-policy}. As a result, we obtained a checkpoint that is slightly more aggressively quantized than its AutoQuantize counterpart, and which performed similarly in terms of accuracy. As with the Nemotron-3-Super NVFP4 checkpoint, we store the Mamba SSM state in FP16 precision using stochastic rounding to reduce runtime costs while preserving accuracy.

\begin{table}[t]
\centering
\small
\begin{tabular}{lccc}
\toprule
Component & BF16 Baseline & FP8 Checkpoint & NVFP4 Checkpoint \\
\midrule
Embedding & BF16 & BF16 & BF16 \\
\midrule
Attention QKV / output projection GEMMs & BF16 & BF16 & BF16 \\
KV cache & FP8 & FP8 & FP8 \\
\midrule
Sparse and shared MoE GEMMs & BF16 & FP8 & NVFP4 \\
MoE latent projection GEMMs & BF16 & BF16 & BF16 \\
Router & FP32 & FP32 & FP32 \\
\midrule
Mamba GEMMs & BF16 & FP8 & FP8 \\
Mamba SSM cache & FP32 & FP32 & FP16+SR \\
Mamba 1D convolution & BF16 & BF16 & BF16 \\
\midrule
LM head & BF16 & BF16 & BF16 \\
\bottomrule
\end{tabular}
\caption{Operator quantization precision policy for the Puzzle-75B-A9B FP8 and NVFP4 checkpoints.}
\label{tab:superturbo-quant-policy}
\end{table}

\subsection{MTP and Speculative Decoding}
\label{sec:mtp_specdec}

As in Nemotron-3-Super, Puzzle-75B-A9B inherits native speculative-decoding support through a Multi-Token Prediction (MTP) head. In this setup, the model contains internal draft layers that predict multiple future tokens from the target model's hidden states. During inference, these draft tokens are verified by the target model allowing multiple tokens to be accepted in a single verification step. Tokens are accepted when the target agrees with the draft distribution either via rejection sampling like in ~\cite{specdec} or naive 1-1 token matching. This provides a lightweight alternative to external draft models: the drafter is part of the same checkpoint, shares the tokenizer and representation space, and introduces only modest additional compute.

Standard MTP implementations often train multiple independent heads, where each head is assigned a fixed prediction offset. For example, one head predicts the next token, the second predicts the token after that, and so on. This methodology limits the draft length during inference to the number of trained heads. To generate longer drafts, one must train more heads, which increases parameters and training complexity. Similarly to Nemotron-3-Super, Puzzle-75B-A9B uses an alternative shared MTP-head formulation: parameters are shared across MTP steps during training, yielding a single prediction head that is exposed to multiple offsets and can be applied recursively at inference time.

The key unaddressed challenge in Super is that even with a shared head, naive teacher-forced MTP training does not match autoregressive MTP inference. Let the input to MTP step $n$ be $(h_1, \ldots, h_n)$, and let the output hidden states be $(h'_2, \ldots, h'_{n+1})$. During training, the $n+1$ MTP step receives the full shifted sequence $(h'_2, \ldots, h'_{n+1})$. At inference time, however, autoregressive drafting changes the conditioning structure. The newly produced state $h'_{n+1}$ is generated while attending to the previous target-model hidden states $(h_1, \ldots, h_n)$. Therefore, the effective input in step $n+1$ becomes $(h_1, \ldots, h_n, h'_{n+1})$, rather than a sequence entirely produced by the MTP module, as seen during training. Continuing one more step, the effective input in step $n+2$ becomes $(h_1, \ldots, h_n, h'_{n+1}, h''_{n+2})$. Thus, as draft length grows, later MTP steps condition on a mixture of target-model hidden states and increasingly noisy MTP-generated hidden states. This distribution differs from the teacher-forced training distribution and can reduce acceptance rates at deeper draft positions.

For Puzzle-75B-A9B we first evaluated a simple transfer recipe: applying the MTP head trained for Nemotron-3-Super directly on top of the Puzzle-75B-A9B backbone. Since Puzzle-75B-A9B is derived from Super, this tests whether the compressed and recovered model remains compatible with the original MTP representation space. We observed similar acceptance lengths, indicating that the Super MTP head transfers well to Puzzle-75B-A9B.
To boost MTP accuracy in Puzzle-75B-A9B further, we continued training the transferred MTP head with a training scheme designed to make the shared MTP head more robust to the training-inference mismatch described above. This stage does not introduce additional MTP heads or any additional runtime mechanisms. Instead, it improves the stability of the internal MTP drafter under autoregressive use, especially at later draft positions where mismatch effects are most visible. The benefit of this training scheme is highlighted by the 25\%-30\% average acceptance length improvement shown in Table~\ref{tab:super_turbo_mtp_speedbench}.

\begin{table}[t]
\centering
\caption{Results of Super and  Puzzle-75B-A9B across BF16 and NVFP4 precisions}
\begin{tabular*}{\linewidth}{@{\extracolsep{\fill}}l|cccc@{}}
\toprule
Benchmark & \shortstack{Puzzle-75B-A9B\\NVFP4} & \shortstack{Super\\NVFP4} & \shortstack{Puzzle-75B-A9B\\BF16} & \shortstack{Super\\BF16} \\
\midrule
\rowcolor{gray!12}\multicolumn{5}{@{}l@{}}{\textbf{General Knowledge}} \\
MMLU-Pro & 82.2 & 83.5 & 82.4 & 83.8 \\
\midrule
\rowcolor{gray!12}\multicolumn{5}{@{}l@{}}{\textbf{Reasoning}} \\
AIME25 (no tools) & 89.9 & 89.9 & 89.7 & 92.2 \\
HMMT Feb25 (no tools) & 92.9 & 93.7 & 93.4 & 94.2 \\
HMMT Feb25 (with tools) & 93.1 & 94.7 & 93.9 & 95.5 \\
GPQA (no tools) & 78.0 & 79.7 & 78.6 & 80.5 \\
GPQA (with tools) & 78.2 & 81.4 & 79.5 & 81.3 \\
LiveCodeBench (v5 2024-07$\leftrightarrow$2024-12) & 79.9 & 81.5 & 81.1 & 82.1 \\
SciCode (subtask) & 40.3 & 41.5 & 40.6 & 42.3 \\
HLE (no tools) & 15.7 & 17.9 & 16.5 & 18.5 \\
\midrule
\rowcolor{gray!12}\multicolumn{5}{@{}l@{}}{\textbf{Agentic}} \\
Terminal Bench (hard subset) & 23.4 & 24.9 & 24.0 & 25.5 \\
SWE-Bench (OpenHands) & 56.9 & 58.7 & 56.9 & 59.5 \\
\textbf{TauBench V2} &  &  &  &  \\
\quad Airline & 55.7 & 55.1 & 55.8 & 56.9 \\
\quad Retail & 63.7 & 64.9 & 63.2 & 64.3 \\
\quad Telecom & 60.3 & 63.8 & 61.5 & 61.1 \\
\quad Average & 59.9 & 61.3 & 60.2 & 60.8 \\
\midrule
\rowcolor{gray!12}\multicolumn{5}{@{}l@{}}{\textbf{Chat \& Instruction Following}} \\
IFBench (prompt) & 71.3 & 71.4 & 71.9 & 73.4 \\
Scale AI Multi-Challenge & 55.9 & 57.3 & 56.6 & 56.6 \\
Arena-Hard-V2 & 69.0 & 72.6 & 68.6 & 72.8 \\
\midrule
\rowcolor{gray!12}\multicolumn{5}{@{}l@{}}{\textbf{Long Context}} \\
AA-LCR & 57.1 & 57.3 & 56.9 & 56.8 \\
RULER 256k & 95.3 & 96.1 & 95.1 & 96.7 \\
RULER 512k & 94.8 & 95.3 & 94.2 & 95.7 \\
RULER 1M & 93.2 & 93.8 & 92.2 & 93.9 \\
\midrule
\rowcolor{gray!12}\multicolumn{5}{@{}l@{}}{\textbf{Multilingual}} \\
MMLU-ProX (avg over langs) & 76.5 & 79.0 & 77.5 & 79.5 \\
WMT24++ (en$\to$xx) & 85.1 & 86.6 & 85.2 & 86.8 \\
\bottomrule
\end{tabular*}
\label{tab:super_main_results}
\end{table}

\section{Experimental Setup and Evaluations}

\subsection{Accuracy Benchmarks}
\label{sec:accuracy_benchmarks}
Table \ref{tab:super_main_results} presents the main benchmark comparison between Nemotron-3-Puzzle-75B-A9B and the original Nemotron-3-Super model across a diverse suite of evaluations spanning reasoning, coding, long-context understanding, multilinguality, instruction following, and agentic behavior. Overall, Puzzle-75B-A9B preserves most of the parent model’s capability despite substantial architectural compression and deployment-oriented optimization, while enabling significantly improved inference efficiency.

On reasoning and coding benchmarks, Puzzle-75B-A9B remains highly competitive with the teacher model,

 maintaining near-teacher performance on HMMT, GPQA, LiveCodeBench, AIME25 and SciCode. Importantly, these capabilities remain robust even under aggressive NVFP4 quantization, where performance degradation is generally modest. These results suggest that the combination of Iterative Puzzle, large-scale KD, and RL recovery

 effectively preserves the model’s core reasoning and software-engineering capabilities.

Puzzle-75B-A9B also demonstrates strong long-context performance despite its reduced model capacity.

Across the RULER evaluations at 256K, 512K, and 1M context lengths, the compressed model remains within roughly 1--2 points of the parent model, while the NVFP4 checkpoint closely tracks BF16 performance. Similar trends are observed on multilingual evaluations such as MMLU-ProX and WMT24++, indicating that the compression pipeline preserves broad cross-lingual capabilities.

The largest performance gaps appear on some instruction-following and agentic evaluations, particularly Arena-Hard-V2 and specific TauBench domains, which appear more sensitive to aggressive compression and low-precision deployment. Nevertheless, the overall results demonstrate that large hybrid MoE models can be substantially compressed while retaining strong downstream performance across a broad range of capabilities, yielding a favorable accuracy--efficiency tradeoff for production inference workloads.

Table~\ref{tab:verbosity_adjusted_throughput} compares the accuracy--throughput tradeoff of Nemotron-3-Super, Nemotron-Labs-3-Puzzle-75B-A9B, and Nemotron-3-Nano under a fixed operating point of 100 user tokens per second. Without speculative decoding, Puzzle-75B-A9B achieves a 2.18× throughput improvement over Super while preserving nearly the same benchmark accuracy (70.74\% vs.\ 71.93\%). Enabling Multi-Token Prediction (MTP) further increases throughput to 4.85× that of Super, while Super's throughput increases to 3.57× after adding MTP. Compared to Nemotron-3-Nano, which achieves the highest raw throughput, Puzzle-75B-A9B+MTP has much higher accuracy while reaching a comparable effective request completion rate after accounting for generation verbosity. In particular, although Nano achieves 7.94× the raw throughput of Super, its higher verbosity reduces its relative request completion rate to 5.26×, compared to 5.14× for Puzzle-75B-A9B+MTP. These results highlight that deployment efficiency should be evaluated not only through token throughput, but also through effective request-level serving performance under realistic generation behavior.

\begin{table}[H]
\centering
\small
\begin{tabular}{llcccccc}
\toprule
Model & Variant & Draft Length & Acc. (\%) & Verbosity & TPS & $\times$ Super & Rel. Req./min \\
\midrule
Super & Single-step & -- & 71.93 & 1.06 & 20{,}939 & 1.00$\times$ & 1.00$\times$ \\
Super & MTP best-DL & 3 & 71.93 & 1.06 & 63{,}604 & 3.04$\times$ & 3.04$\times$ \\
Puzzle-75B-A9B & Single-step & -- & 70.74 & 1.00 & 42{,}601 & 2.03$\times$ & 2.16$\times$ \\
Puzzle-75B-A9B & MTP best-DL & 3 & 70.74 & 1.00 & 96{,}997 & 4.63$\times$ & 4.91$\times$ \\
Nano & Single-step & -- & 60.55 & 1.60 & 122{,}308 & 5.84$\times$ & 3.87$\times$ \\
\bottomrule
\label{tab:speed_frontier}
\end{tabular}

\caption{
Comparison of throughput and effective request completion rate across Nemotron-3-Super, Puzzle-75B-A9B, and Nano. 
TPS denotes server throughput measured at a fixed operating point of 100 user tokens per second. 
``Rel. Req./min'' adjusts throughput by generation verbosity to estimate the relative number of user requests completed per minute by the server.
Draft length denotes the speculative decoding draft length that provides the best speedup, and ``MTP best-DL'' refers to the best-performing Multi-Token Prediction configuration for each model.
}
\label{tab:verbosity_adjusted_throughput}
\end{table}

\begin{figure}[h]
\centering

\begin{minipage}{0.45\textwidth}
\centering
\setlength{\tabcolsep}{3pt}
\resizebox{\linewidth}{!}{
\begin{tabular}{lcc}
\toprule
Category
& Super MTP
& Puzzle-75B-A9B MTP \\
\midrule
Coding        & 3.78 (3.63) & 4.80 (4.79) \\
Humanities    & 3.26 (3.19) & 4.14 (4.12) \\
Math          & 3.73 (3.64) & 4.75 (4.75) \\
Multilingual  & 4.05 (3.76) & 5.31 (5.23) \\
QA            & 3.16 (3.05) & 3.95 (3.93) \\
RAG           & 3.78 (3.63) & 4.78 (4.77) \\
Reasoning     & 3.59 (3.45) & 4.46 (4.44) \\
Roleplay      & 2.82 (2.65) & 3.45 (3.46) \\
STEM          & 3.30 (3.24) & 4.15 (4.16) \\
Summarization & 3.48 (3.22) & 4.22 (4.20) \\
Writing       & 2.99 (2.96) & 3.73 (3.61) \\
\midrule
Average       & 3.45 (3.31) & 4.34 (4.31) \\
\bottomrule
\end{tabular}}
\captionof{table}{MTP average acceptance lengths on the SPEED-Bench \textit{qualitative} split using a draft length of 7. Values in parentheses report the corresponding NVFP4 checkpoint.}
\label{tab:super_turbo_mtp_speedbench}
\end{minipage}
\hfill
\begin{minipage}{0.45\textwidth}  
\centering
\includegraphics[width=\linewidth]{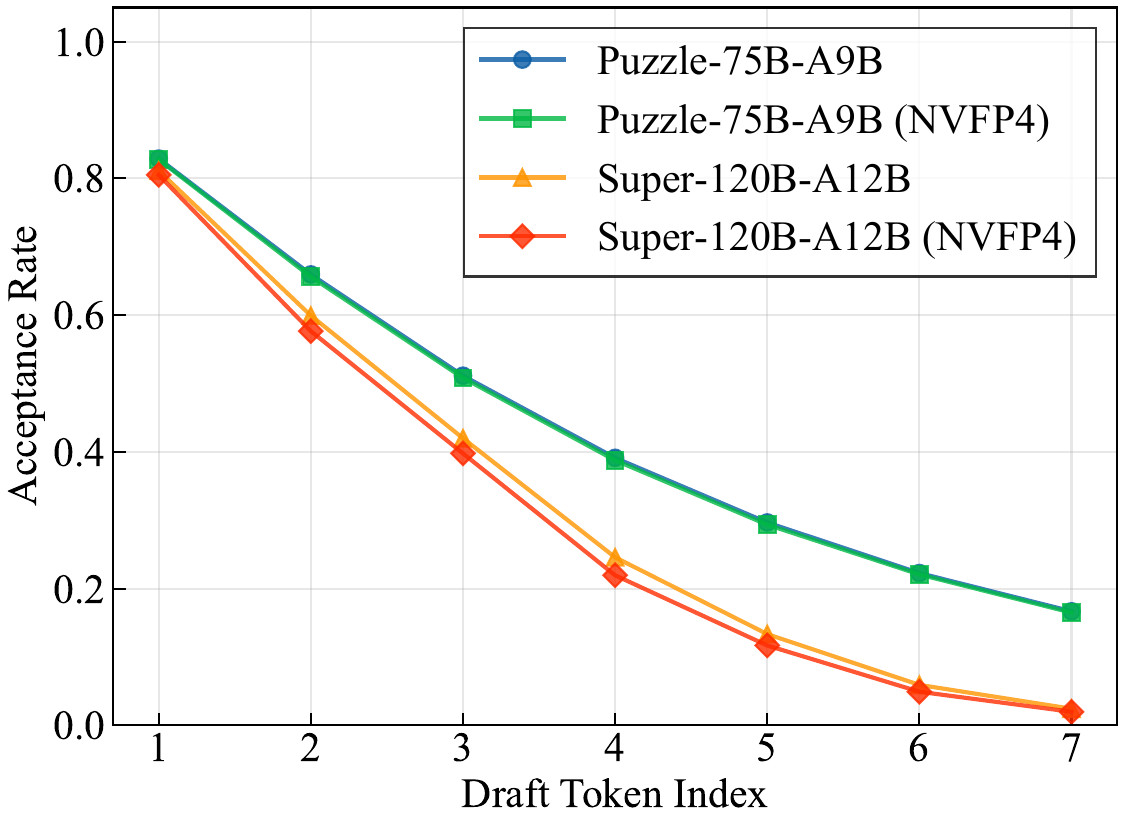}
\caption{MTP acceptance rate by draft token index on the SPEED-Bench \textit{qualitative} split using a draft length of 7.}
\label{fig:super_turbo_mtp_ar_by_draft_index}
\end{minipage}

\end{figure}

\textbf{MTP results}- 
We evaluate MTP quality using SPEED-Bench~\cite{speed_bench}, measuring the per-sample acceptance lengths (ALs) across the categories in the \textit{qualitative} data split. Table~\ref{tab:super_turbo_mtp_speedbench} reports the average ALs by category, while Figure~\ref{fig:super_turbo_mtp_ar_by_draft_index} shows average acceptance rates by draft-token index. The final Puzzle-75B-A9B MTP reaches strong ALs across BF16 and quantized checkpoints, with similar ALs for BF16 and NVFP4. 
This behavior differs from Nemotron-3-Super, where the NVFP4 checkpoint showed a larger degradation in MTP acceptance length relative to BF16. In addition, the continued MTP training described in Section~\ref{sec:mtp_specdec} improves average acceptance length by approximately 25\%-30\% compared to Super, with the gains most visible at deeper draft positions.

\subsection{Ablations}
\label{sec:ablations}
\begin{table}[t]
\centering
\scriptsize
\setlength{\tabcolsep}{4pt}

\begin{tabular}{lcccccc}
\toprule
Method & MMLU-Pro & GPQA & HLE & AA-LCR  & LiveCode & Avg. \\
\midrule
1-step Puzzle & 81.4 & 75.9 & 15.9 & 55.3 & 78.2 & 68.48 \\
3-step Iterative Puzzle & \textbf{82.1} & \textbf{77.4} & \textbf{16.2} & \textbf{57.5} & \textbf{78.9} & \textbf{69.05} \\
\midrule
$\Delta$ & $+0.7$ & $+1.5$ & $+0.3$ & $+2.2$ & $+0.7$ & $+0.57$ \\
\bottomrule
\end{tabular}

\vspace{0.5em}

\begin{tabular}{lccccc}
\toprule
Method & SciCode & RULER-128K & RULER-256K & IFBench-Inst. & IFBench-Prompt \\
\midrule
1-step Puzzle & 41.0 & \textbf{96.4} & 94.5 & \textbf{74.4} & \textbf{71.8} \\
3-step Iterative Puzzle & \textbf{41.1} & \textbf{96.4} & \textbf{95.4} & 74.2 & 71.3 \\
\midrule
$\Delta$ & $+0.1$ & $+0.0$ & $+0.9$ & $-0.2$ & $-0.5$ \\
\bottomrule
\end{tabular}

\caption{
Ablation comparing a single-step Puzzle baseline against the full three-stage Iterative Puzzle procedure. Metrics are reported as percentages or benchmark scores where higher is better. The average is an unweighted mean over all displayed benchmarks. Iterative Puzzle improves the average by $0.57$ points while preserving long-context performance and improving most core benchmarks.}
\label{tab:iterative-puzzle-ablation}
\end{table}
\paragraph{Single-shot vs. iterative compression.}
To isolate the effect of iterative compression, we compare the full three-stage Iterative Puzzle procedure against a single-step Puzzle baseline. The single-step baseline applies the compression targets in one Puzzle run, whereas Iterative Puzzle reaches the final architecture through repeated compression-and-recovery stages with score recomputation between stages. Table~\ref{tab:iterative-puzzle-ablation} shows that the iterative procedure improves the unweighted benchmark average by $0.57$ points, with gains on MMLU-Pro, GPQA, HLE, AA-LCR, LiveCodeBench, SciCode, and RULER-256K. This suggests that recomputing the Puzzle objective after intermediate recovery produces a better final architecture than applying the compression in a single shot.


\paragraph{Training progression.}

\begin{figure}[h]
\centering
\includegraphics[width=0.9\textwidth]{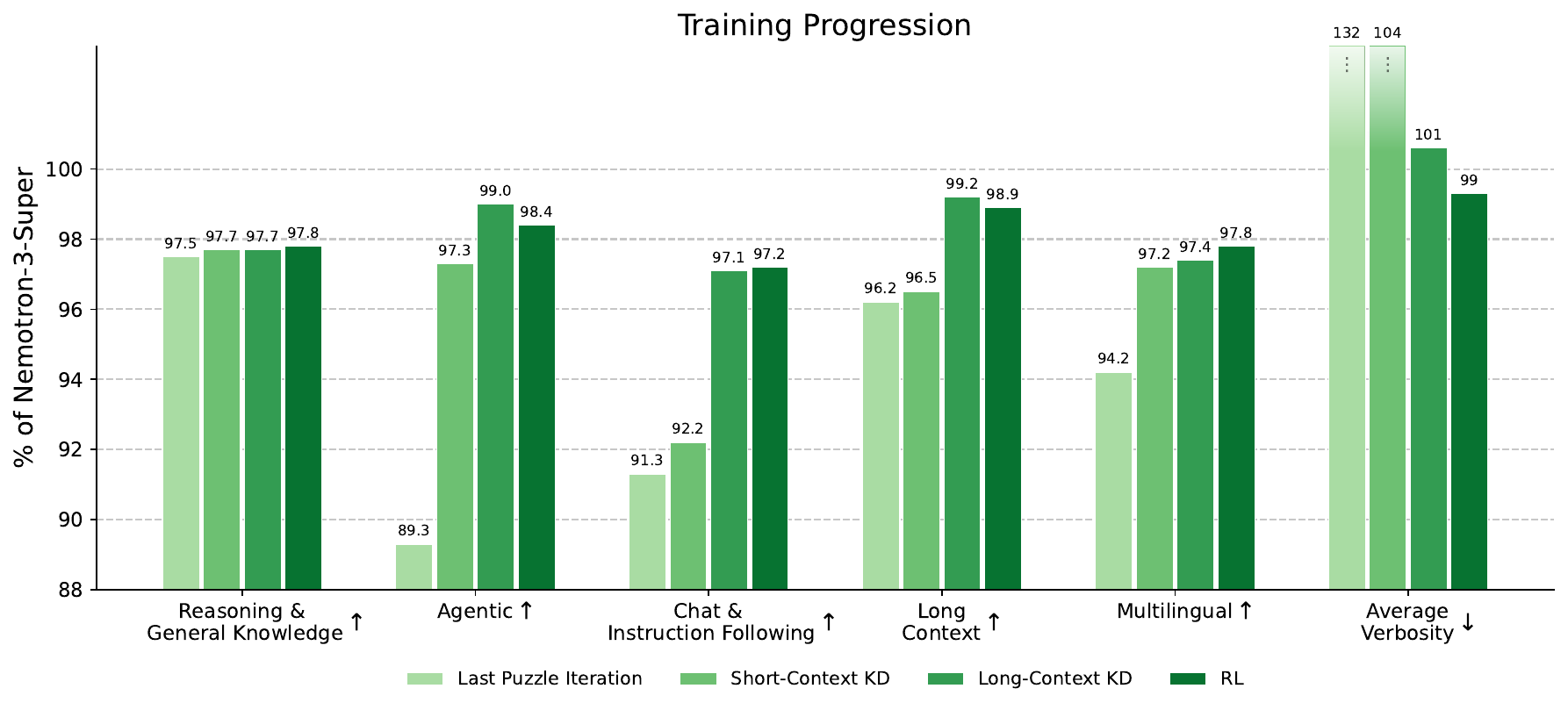}
\caption{Training progression of Puzzle-75B-A9B. Accuracy recovers sharply after the final Puzzle iteration with short-context KD, while long-context KD selectively boosts long context capabilities. The impact of RL was small. Verbosity improves progressively throughout the training stages.}
\label{fig:progression}
\end{figure}

Figure~\ref{fig:progression} shows various aspects of model quality throughout the different training stages of Nemotron-Labs-3-Puzzle-75B-A9B. The training progression highlights the complementary roles of short-context KD, long-context KD, and RL in the recovery phase. The first short-context KD stage is essential for recovering general accuracy after the final Puzzle iteration, bringing the accuracy across most benchmark categories to over 97\% of Nemotron-3-Super. Long-context KD then provides a targeted improvement in benchmarks that contain long input sequences or require long generations. Finally, the impact of RL training in our experiments was small -- we plan to research more RL environments and additional post-training techniques in future work. We note that,  like \cite{gpt-oss-puzzle} we see an increase of verbosity in the compressed model before recovery, but it improves steadily across the training stages, finally reaching parity with the parent model. Benchmark categories match Table~\ref{tab:super_main_results}.

\paragraph{Disaggregated prefill pruning.}
We also study whether prefill--decode disaggregation enables an additional,
deployment-specific compression axis. In this setting, the decode model remains
the full Puzzle-75B-A9B checkpoint, while a smaller prefill-only model is used to
process the prompt and produce the KV cache and Mamba SSM state consumed by the
decode model. As detailed in Appendix~\ref{disagg_prefill_appendix}, we evaluate two prefill-pruned variants
that reduce the prefill MoE top-$k$ and the model's embedding dimension, and train the
prefill and decode models jointly to maintain state compatibility across the
prefill--decode boundary. Both variants preserve most benchmark quality while
improving the prefill-heavy 50K/1K serving throughput by 5--7\% relative to the
same-model Puzzle-75B-A9B baseline. Importantly, using the smaller
prefill model for both prefill and decode substantially degrades the common
benchmark average, indicating that the gain is not simply due to replacing the
full model with a smaller network. Instead, the result supports a targeted
disaggregation strategy: compress the compute-heavy prefill path while retaining
the full decode model for generation quality.

\begin{figure*}[h]
\centering
\includegraphics[width=\linewidth]{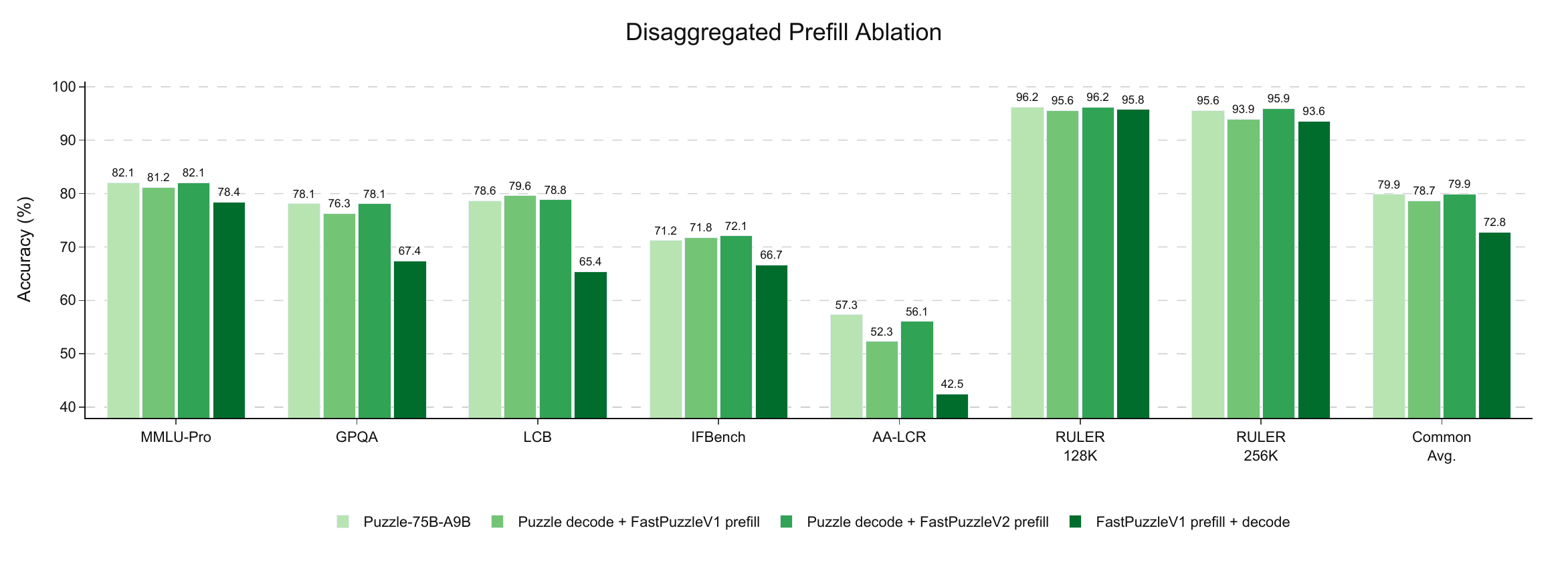}
\caption{Disaggregated prefill ablation. \textsc{FastPuzzleV1} and \textsc{FastPuzzleV2} are faster derivatives
of Puzzle-75B-A9B that are pruned further to accelerate the prefill stage. Using a faster model only for prefill barely
changes accuracy as long as Puzzle-75B-A9B is still used for decode, whereas using the faster model for both
prefill and decode substantially degrades accuracy. Numbers above bars show the exact plotted values.}
\label{fig:disagg-prefill-ablation}
\end{figure*}

\subsection{Performance Analysis}
\label{sec:bench}

We evaluate the inference efficiency of \textbf{Puzzle-75B-A9B} against its parent \textbf{Super} on the two deployment scenarios:

The prefill-heavy 50K\,/\,2K regime and the decode-heavy 8K\,/\,64K regime. We first isolate the gain from architectural compression alone by reporting \emph{single-step} (non-MTP) results, then layer on multi-token prediction gains in the appendix \ref{sec:perf_appendix}.

\paragraph{Hardware and quantization.}
We evaluate three deployment settings, each with the parent and the compressed model served at matched quantization so that the measured gap reflects compression alone, not a change in numeric format:
\begin{itemize}
  \item \textbf{Single 8$\times$B200 node (primary target)} --- NVFP4 weights, FP8 KV cache, FP16 Mamba SSM state with stochastic rounding.
  \item \textbf{Single 8$\times$H100 node} --- FP8 weights, FP8 KV cache, FP32 Mamba SSM state. Not a primary optimization target, but reported for completeness. 
  \item \textbf{Single H100 GPU (1M-context target)} --- NVFP4 weights, FP8 KV cache, FP32 Mamba SSM state.  NVFP4 is not natively supported on Hopper, but it is beneficial here because HBM capacity is the binding constraint at 1M context.
\end{itemize}
We present the results of the 8-B200 in the main text and the two other scenarios in the appendix \ref{sec:perf_appendix}.
\paragraph{Scenarios and notation.}
Inference scenarios are denoted as \emph{input/output} token lengths. We use \emph{user throughput} (UT, tokens per user per second) for the per-request generation rate and \emph{total throughput} (TPS, tokens per second) for the aggregate rate across all concurrent users on the node.

\paragraph{Pareto methodology.}
For each (model, scenario) cell we sweep tensor-parallel degree TP\,$\in\{1,2,4,8\}$, with both EP-on and EP-off variants, and a batch-size grid from 1 up to the maximum batch size that fits in memory for that parallelization strategy. Each configuration is repeated twice. We then compute the Pareto frontier on the (UT, TPS) plane: at each user-throughput level, the Pareto-optimal point is the configuration that maximizes total throughput. For each user-throughput threshold $\tau$, we report the highest TPS along the Pareto frontier subject to UT\,$= \tau$. This formulation directly answers the deployment question, ``how much can a serving system push through under an interactivity constraint of $\tau$ tokens per user per second?''

\begin{figure}[H]
\centering
\includegraphics[width=1\textwidth]{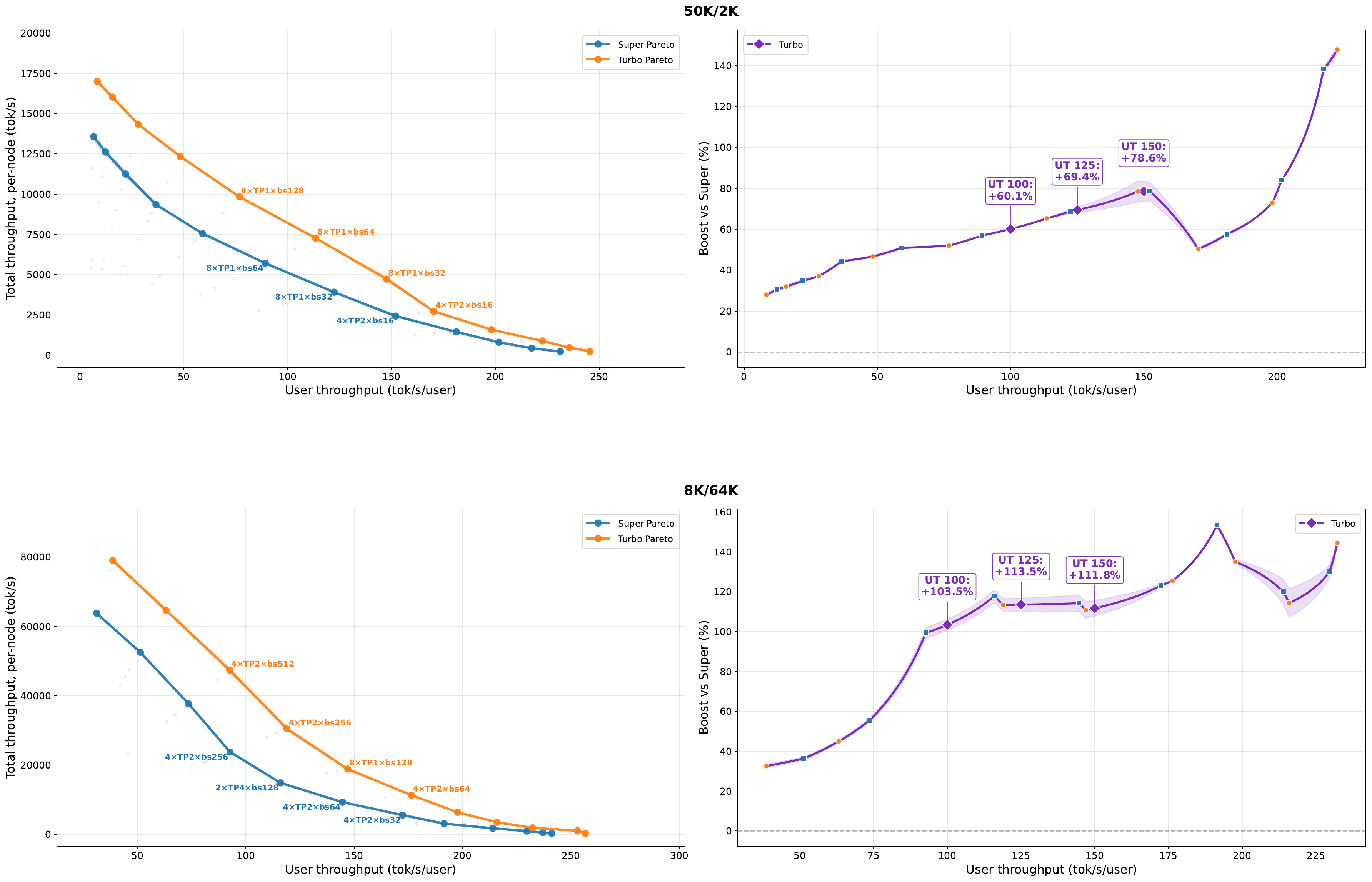}
\caption{Pareto frontiers underlying Table~\ref{tab:pareto-single-step}, on the 8$\times$B200 node with NVFP4 weights / FP8 KV / FP16 Mamba state. Each curve is the upper envelope of total vs.\ user throughput across all (TP, EP, batch size) configurations. Dashed verticals mark the UT thresholds in Table~\ref{tab:pareto-single-step}. The shaded regions represent variability across measurements.}
\label{fig:pareto-single-step}
\end{figure}

\paragraph{8$\times$B200 node, interactive serving.}
Table~\ref{tab:pareto-single-step} reports the Pareto-optimal total throughput for both models at three increasingly stringent user-throughput thresholds, spanning the interactive regime: UT$=100$, UT$=125$, and UT$=150$. Across all six operating points, single-step Puzzle-75B-A9B delivers between $1.6\times$ and $2.14\times$ the total throughput of Super. Figure~\ref{fig:pareto-single-step} shows the full Pareto curves underlying the table.

\begin{table}[H]
\centering
\small
\begin{tabular}{@{}llrrr@{}}
\toprule
Scenario & UT operating point & Super & Puzzle-75B-A9B & Boost \\
         &                    & (tok/s) & (tok/s)   &       \\
\midrule
50K\,/\,2K & $\geq 100$ &  5{,}128 &  8{,}210 & $1.60\times$ \\
           & $\geq 125$ &  3{,}784 &  6{,}412 & $1.69\times$ \\
           & $\geq 150$ &  2{,}532 &  4{,}523 & $1.79\times$ \\
\addlinespace
8K\,/\,64K & $\geq 100$ & 20{,}939 & 42{,}601 & $2.03\times$ \\
           & $\geq 125$ & 13{,}074 & 27{,}918 & $2.14\times$ \\
           & $\geq 150$ &  8{,}522 & 18{,}047 & $2.12\times$ \\
\bottomrule
\end{tabular}

\caption{Pareto-optimal total throughput at three interactive
user-throughput operating points, single-step decoding (no MTP). For
each cell we linearly interpolate the (TP, EP, batch size) Pareto
envelope at exactly the listed UT value. Both models served at
matched NVFP4 weights, FP8 KV cache, and FP16 Mamba SSM state with
stochastic rounding, on a single 8$\times$B200 node. ``Boost'' is
Puzzle-75B-A9B $/$ Super at matched UT. Our design target was a
$1.6\times$ boost at UT$=100$ tok/s; the achieved boost ranges from
$1.6\times$ in the prefill-heavy 50K/2K regime to $2.14\times$ in
the decode-heavy 8K/64K regime at the most stringent UT$=150$ tok/s
operating point.}
\label{tab:pareto-single-step}
\end{table}

\section{Conclusions}
In this work, we presented Nemotron-Labs-3-Puzzle-75B-A9B, a deployment-optimized compressed variant of Nemotron-3-Super designed for high-throughput inference workloads. Our approach combines Iterative Puzzle, a sequential extension of the Puzzle framework for hardware-aware architectural compression, with large-scale knowledge distillation, reinforcement learning recovery, quantization, and speculative decoding techniques. The resulting model substantially reduces both active and total model parameters

while preserving strong accuracy across reasoning, coding, multilingual, long-context, and agentic evaluations.

Our results demonstrate that large hybrid MoE models can be substantially compressed while maintaining strong downstream capabilities and significantly improving deployment efficiency. In interactive serving settings, Puzzle-75B-A9B achieves roughly 2× higher throughput than the parent model at matched user-throughput constraints, while in 1M-token single-GPU deployments, it increases sustainable concurrency from 1 to 8 requests.

More broadly, this work highlights the effectiveness of iterative, hardware-aware compression as a practical path toward scalable deployment of frontier-scale LLMs across diverse inference environments.
In future work, we plan to use Iterative Puzzle more aggressively, together with expanding our current pruning techniques, to enable extreme LLM compression -- reaching much higher compression rates while preserving strong model accuracy.

\clearpage
\bibliography{ref}
\bibliographystyle{plainnat}

\clearpage
\appendix
\section*{Appendix}
\section{Mamba SSM Pruning}
\label{appendix:ssm_pruning}
\subsection{Method}
We chose which SSM channels to prune by estimating their contribution to the Mamba layer output in the following manner. 

Given an input vector $x_t$, the SSM layer computes the following transformation:

\[
\begin{aligned}
h_{t}^{(i)} &= A_t h_{t-1}^{(i)} + B_t^{(i)} x_t \\
y_t^{(i)} &= {C_t^{(i)}}^{\top} h_t^{(i)} \\
y_t &= \sum_{i=1}^{N} y_t^{(i)}
\end{aligned}
\]

$h_t^{(i)}$ denotes the latent state at time step $t$ for SSM channel $i \in \{1, ..., N\}$. $A_t$, $B_t$, and $C_t$ are the SSM matrices at time step $t$. $y_t$ is the SSM output. The shape of $x_t$ and $y_t$ is $H \cdot P$, where $H$ is the number of Mamba heads and $P$ is the head dimension. The heads are divided into $G$ groups with $H/G$ heads each.

After the SSM stage, the Mamba-2 computation proceeds as follows:
\[
\begin{aligned}
\mathrm{out}_t &= W_o \, \mathrm{Norm}\!\left(y_t \odot g_t\right)
\end{aligned}
\]

$\mathrm{out}_t$ is the final output of the Mamba layer, given by multiplying $y_t$  by a gate vector $g_t$,
applying a grouped LayerNorm, and finally applying the output projection $W_o$. 

To calculate SSM channel contribution, we postpone the channel sum to the latest truly decomposable step, right before the grouped LayerNorm:
\[
\begin{aligned}
y_t \odot g_t
&= \sum_{i=1}^{N} \left(y_t^{(i)} \odot g_t\right) \coloneqq \sum_{i=1}^{N} \mathrm{contrib}_t^{(i)}
\end{aligned}
\]

$\mathrm{contrib}_t^{(i)}$ can be separated into the different Mamba groups by reshaping it to $[G, H \cdot P\,/\,G]$. To estimate the importance of channel $i$ in group $g$, we average $ \lvert contrib_t^{(i)} \rvert $ over all the heads in group $g$ for 67M tokens of validation data. We use this metric to rank all the SSM channels inside each group, finally reducing the total number of SSM channels from $G \times N = 8 \times 128$ to $G \times N' = 8 \times 96$.

\subsection{Results}
\begin{figure}[H]
\centering
\includegraphics[width=1.0\textwidth]{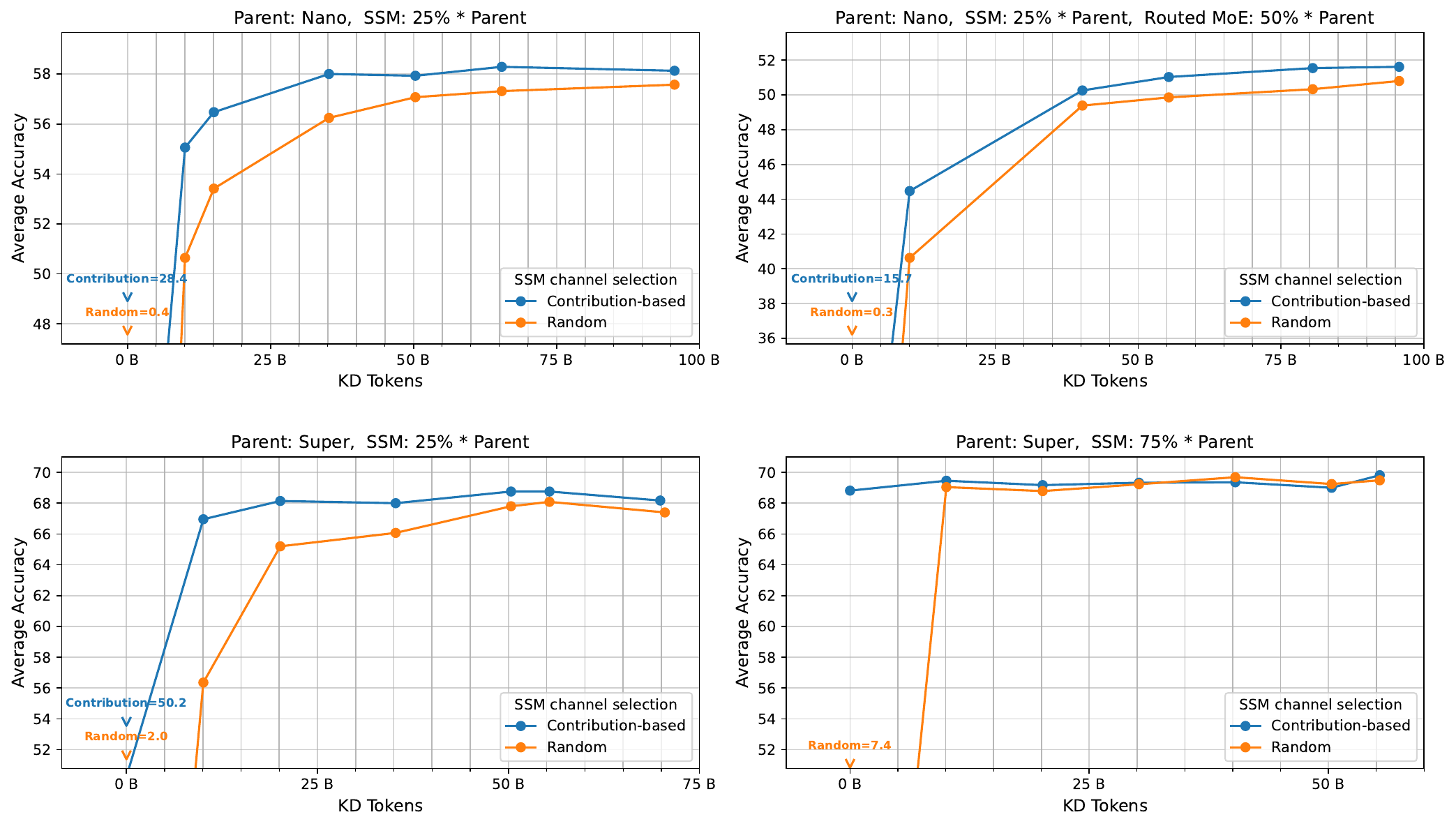}
\caption{The effect of the SSM channel-selection method. Under aggressive pruning, our contribution-based channel selection consistently outperforms random channel selection both before and after KD, while also requiring fewer KD tokens to recover accuracy.}
\label{fig:mamba_pruning}
\end{figure}

Figure~\ref{fig:mamba_pruning} shows the effect of our SSM channel-selection method. Under aggressive pruning, where most of the parent model’s SSM channels are removed, our contribution-based channel selection consistently outperforms random channel selection both before and after KD, while also requiring fewer KD tokens to recover accuracy.

At more moderate pruning rates, such as the setting used for Nemotron-3-Super-Turbo (Table~\ref{tab:final-architecture}), contribution-based selection yields substantially higher initialization accuracy than random selection (68.8 vs. 7.4), although KD closes this gap. Nevertheless, high-accuracy initializations are useful when using Puzzle to build models with heterogeneous SSM channel pruning, to accurately estimate the quality of each Mamba layer in the block library. We leave this for future work, since inference frameworks do not currently support a different SSM state size per layer.

In this section, we used the following benchmarks to estimate accuracy: AA-LCR, GPQA Diamond, HLE, IFBench, LiveCodeBench, MMLU-Pro, RULER-256K.

\section{Disaggregated Prefill Pruning}
\label{disagg_prefill_appendix}
As model sizes continue to grow, prefill--decode disaggregation becomes an increasingly attractive serving
strategy for high-volume, latency-sensitive deployments. In this setting, prompt processing is handled by a
prefill pool, often spread over a separate set of nodes, while autoregressive generation is handled by a decode
pool. The prefill workers process the input sequence in parallel and materialize the attention KV cache and
the Mamba SSM state required by the decode workers. These states are then transferred to the decode pool,
which continues generation token by token. This separation is useful because the two phases stress the system
in different ways: prefill is dominated by large prompt compute, whereas decode is typically more sensitive
to cache/state bandwidth, batching, and per-user token latency. Disaggregating the two phases allows each
pool to be scaled and batched independently, reducing head-of-line blocking from long prompts and keeping
decode resources focused on low-latency generation.

We test whether this deployment structure creates an additional compression opportunity: the prefill model
does not need to be the same network as the decode model, as long as it produces KV caches and Mamba
states that remain compatible with the decode model. Starting from Puzzle-75B-A9B, we construct two faster
prefill-only derivatives, which we denote \textsc{FastPuzzleV1} and \textsc{FastPuzzleV2}. These variants prune
the compute-heavy dimensions most relevant to the prefill phase while keeping the full Puzzle-75B-A9B model
as the decode model. \textsc{FastPuzzleV1} reduces prefill top-$k$ to $20\%$ of the Super value and the
embedding dimension to $93\%$ of the Super embedding dimension. \textsc{FastPuzzleV2} uses a milder
configuration, with prefill top-$k$ reduced to $30\%$ of the Super value and embedding dimension reduced to
$97\%$ of the Super embedding dimension.

To maintain state compatibility, we train the large decode model and the smaller prefill model jointly using
Star Elastic training~\citep{taghibakhshi2026starelastic}. In our disaggregated setting, the states produced by
the prefill model are used directly by the decode model: the KV cache and Mamba SSM state are transferred
across the prefill--decode boundary without any learned projection, adapter, cache translation module, or
recomputation by the decode model. The joint training therefore serves two purposes: it heals the smaller
prefill model and aligns its generated cache/state distribution with the state distribution expected by the full
Puzzle-75B-A9B decode model.

Figure~\ref{fig:disagg-prefill-ablation} evaluates the common benchmark set available across all compared
checkpoints: MMLU-Pro, GPQA, LiveCodeBench, IFBench, AA-LCR, RULER-128K, and RULER-256K, together
with their unweighted common average. The disaggregated variants preserve most of the Puzzle-75B-A9B
quality while improving the prefill-heavy 50K/1 serving point: \textsc{FastPuzzleV1} improves prefill
throughput by $7\%$, and \textsc{FastPuzzleV2} improves it by $5\%$, relative to the Puzzle-75B-A9B prefill
baseline. On the common average, using Puzzle-75B-A9B for decode with \textsc{FastPuzzleV1} or
\textsc{FastPuzzleV2} for prefill gives $78.7$ and $79.9$, respectively, compared with $79.9$ for the
same-model Puzzle-75B-A9B baseline. In contrast, using \textsc{FastPuzzleV1} for both prefill and decode
drops the common average to $72.8$, with especially large degradation on GPQA, LiveCodeBench, IFBench,
and AA-LCR. This supports the hypothesis that disaggregated serving enables a targeted compression point:
make the prefill path cheaper while retaining the full decode model for generation quality.


\section{Performance Analysis}
\label{sec:perf_appendix}
\begin{figure}[H]
\centering
\includegraphics[width=0.92\textwidth]{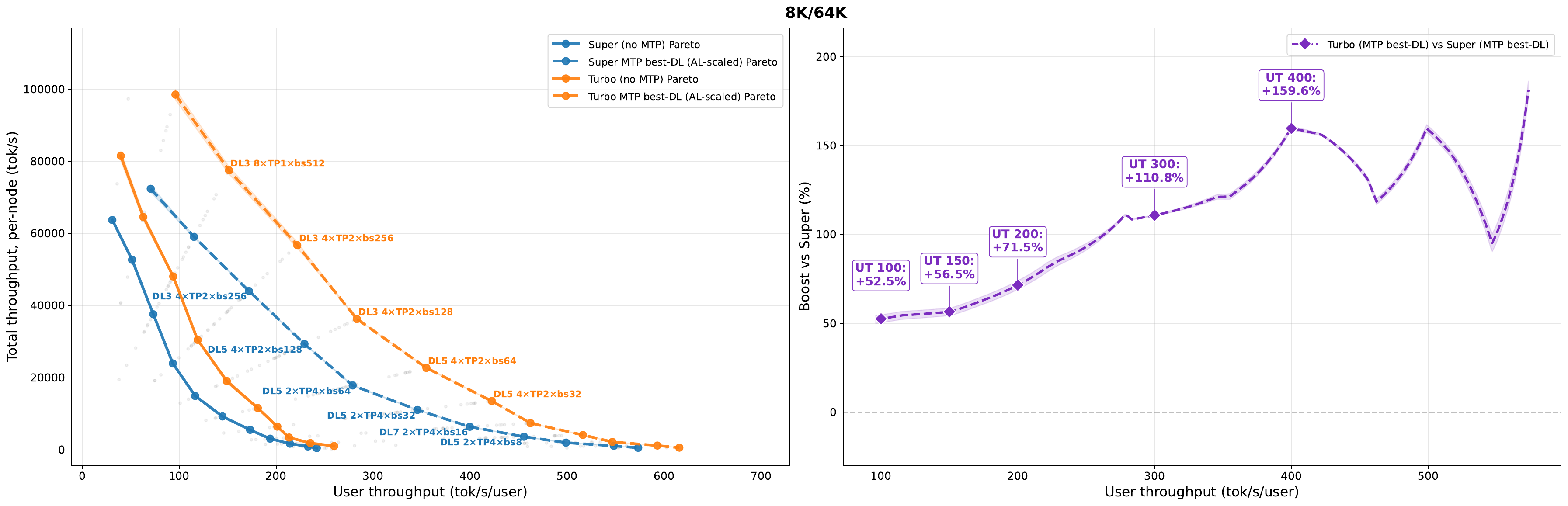}
\caption{Super Turbo Pareto frontier with MTP overlaid on the single-step curves from Figure~\ref{fig:pareto-single-step}, same hardware and quantization. Solid: no MTP. Dashed: MTP best-DL envelope across DL\,$\in\{1,3,5,7\}$.}
\label{fig:pareto-mtp}
\end{figure}

Multi-token prediction (MTP, \S\ref{sec:mtp_specdec}) compounds these gains by emitting up to $\mathrm{DL}$ candidate tokens per forward pass and accepting whichever the verifier confirms; on average each step produces $\mathrm{AL}$ accepted tokens, so per-step throughput scales by approximately $\mathrm{AL}$ at fixed kernel cost. Figure~\ref{fig:pareto-mtp} overlays Super Turbo's MTP Pareto frontier - the upper envelope across DL\,$\in\{1,3,5,7\}$ at the synthetic acceptance lengths reported in \S\ref{sec:mtp_specdec} - on the single-step curves of Figure~\ref{fig:pareto-single-step}. \emph{TODO: drop in the headline MTP boost numbers (Super-Turbo-MTP-best vs.\ Super-Turbo-no-MTP at UT$\geq 100$, both scenarios) from the latest report.}

\begin{figure}[H]
\centering
\includegraphics[width=1\textwidth]{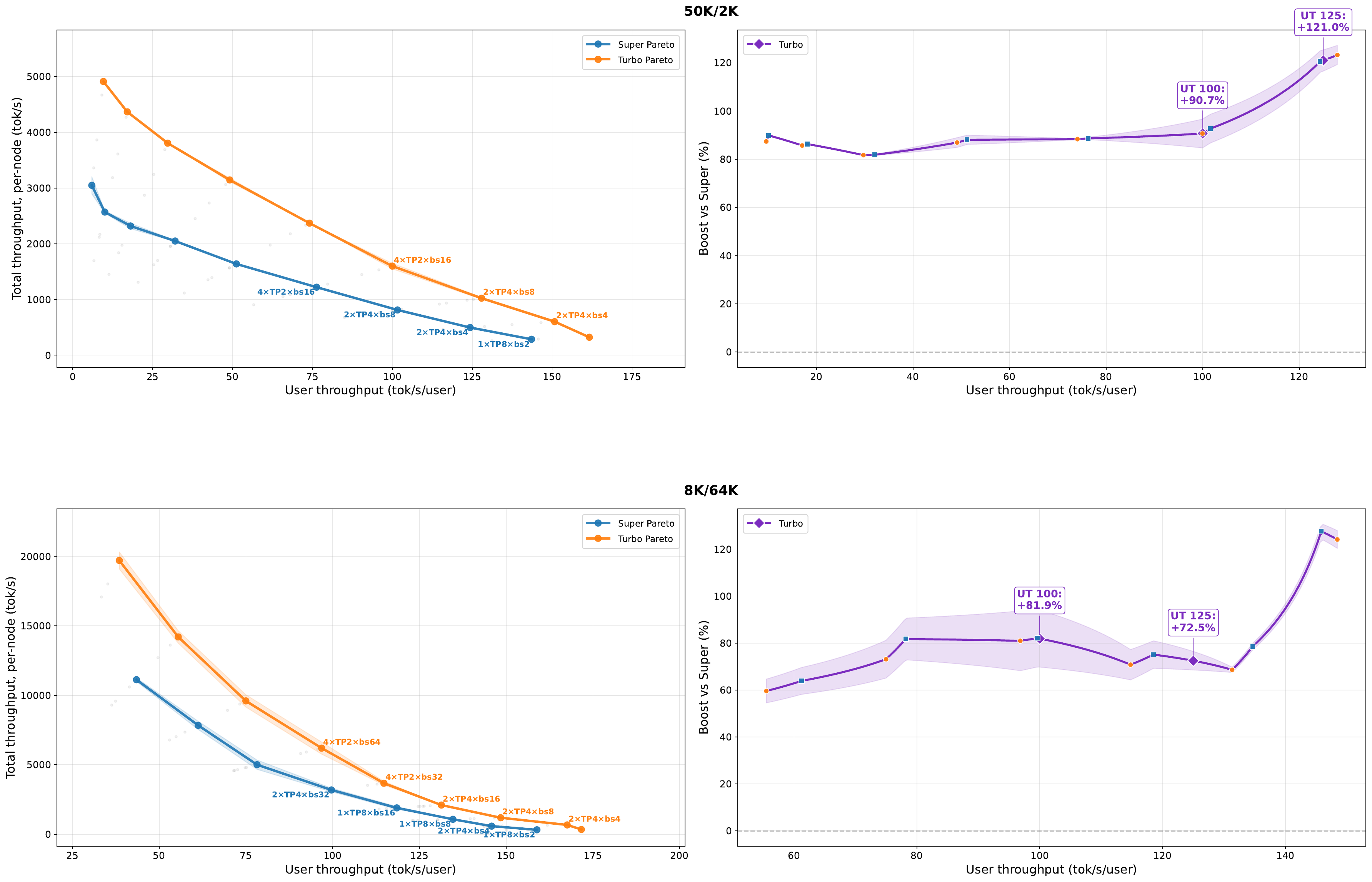}
\caption{Pareto frontiers of total vs.\ user throughput on a single
8$\times$H100 node, FP8 weights / FP8 KV / FP32 Mamba SSM, single-step
decoding (no MTP). Each curve is the upper envelope across all
(TP, EP, batch size) configurations. The dashed vertical at
UT\,$=\!100$ tok/s marks the operating point used in the interpolated
boost reported in the text: Super Turbo delivers $1.91\times$ the
total throughput of Super on 50K\,/\,2K and $1.82\times$ on 8K\,/\,64K
at this user-throughput target.}
\label{fig:pareto-single-step-h100}
\end{figure}

\paragraph{8$\times$H100 node, interactive serving.}
We repeat the Pareto analysis on a single 8$\times$H100 node, with both
models served at matched FP8 weights, FP8 KV cache, and FP32 Mamba SSM
state. Figure~\ref{fig:pareto-single-step-h100} shows the resulting
frontiers. Interpolating each model's Pareto envelope at exactly
UT\,$=\!100$ tok/s gives Super Turbo $1.91\times$ the total throughput
of Super on 50K\,/\,2K (1{,}599 vs.\ 838 tok/s) and $1.82\times$ on
8K\,/\,64K (5{,}753 vs.\ 3{,}163 tok/s). The H100 stack is not a primary
optimization target, but these numbers confirm that the architectural
gains of Super Turbo transfer to the Hopper deployment.

\paragraph{Single H100 at 1M context.}
At 1M-token context on a single H100, the binding constraint flips from compute to memory: the GPU's 80\,GB HBM budget is split between model weights and per-request KV cache (Mamba state is negligible at this scale). Super's weights occupy $\sim$70\,GB at NVFP4, and a single 1M-token request adds $\sim$4\,GB of KV cache; together these saturate the budget, so the parent's effective concurrency at 1M is 1. Super Turbo preserves the parent's attention layout (and therefore the same $\sim$4\,GB-per-request KV cost), but its $\sim$44.5\,GB NVFP4 weight footprint frees enough HBM to host seven additional 1M-token requests, lifting concurrency at 1M to 8.

At this concurrency, Super Turbo delivers approximately $4\times$ the aggregate decode throughput of Super (Super Turbo at bs\,=\,8: $\sim$400 tok/s, derived from 20.1\,ms median inter-token latency across 8 concurrent requests; Super at bs\,=\,1: $\sim$94 tok/s, derived from 10.7\,ms TPOT). Per-request prefill of the 990K-token prompt is roughly $1.2\times$ faster on Super Turbo than on Super.

\newcommand{\latenthlt}[1]{\textcolor[HTML]{50A700}{#1}}

\section{Latent Dimension Pruning Ablation}
\label{sec:latent-dim-ablation}

Another axis of pruning the Super MoE layers is the shared latent (bottleneck) dimension of the MoE block itself. A latent MoE layer~\citep{latentmoe} maps an input hidden state \(x \in \mathbb{R}^{H}\) to
\begin{align*}
	\latenthlt{\mathrm{Latent}}\mathrm{MoE}(x)
	=
	\latenthlt{W_{\mathrm{out}}}
	\left(
	\sum_{k \in \mathcal{T}(x)}
	g_k(x)\,D_k\,\sigma(U_k \latenthlt{W_{\mathrm{in}}} x)
	\right)
\end{align*}
where \(\mathcal{T}(x)\) denotes the set of routed experts selected for token \(x\), \(g_k(x)\) are the corresponding router weights, and \(\sigma(\cdot)\) is the activation function. The layer is parameterized by the shared latent projections
\begin{align*}
	W_{\mathrm{in}} &\in \mathbb{R}^{L \times H}, &
	W_{\mathrm{out}} &\in \mathbb{R}^{H \times L},
\end{align*}
and the expert-specific up/down projections
\begin{align*}
	U_k &\in \mathbb{R}^{M \times L}, &
	D_k &\in \mathbb{R}^{L \times M}.
\end{align*}
The shared projections $W_{\mathrm{in}}$ and $W_{\mathrm{out}}$ map between the model's hidden dimension $H$ and a smaller latent dimension $L \ll H$ in which all experts operate.
We study pruning this bottleneck from $L$ to smaller input and output latent sizes $(L'_{\mathrm{in}},L'_{\mathrm{out}})$. Reducing these dimensions shrinks every routed expert in the layer and also reduces the shared latent projections, making latent pruning structurally complementary to intermediate-channel pruning. We compare coordinate-selection initializers that keep subsets of original latent coordinates against a projection-based method that can mix them.

\paragraph{Channel-selection baselines.} The coordinate-aligned baselines choose a selection set $\mathcal{S}\subset\{1,\dots,L\}$ of size $L'$ for the latent side being pruned. On the input side, this keeps the selected rows of $W_{\mathrm{in}}$ and the corresponding columns of every $U_k$; on the output side, it keeps the selected columns of $W_{\mathrm{out}}$ and the corresponding rows of every $D_k$:
\begin{itemize}
	\item \emph{Minitron}~\citep{minitron}. Activation-based importance: for each latent coordinate $i \in \{1,\dots,L\}$, score $s_i = \mathbb{E}_x |z_i|$ on the input side (with $z = W_{\mathrm{in}} x$ the post-latent-projection activation) and analogously $s_i = \mathbb{E}|z^{\mathrm{out}}_i|$ on the output side (with $z^{\mathrm{out}}$ the pre-$W_{\mathrm{out}}$ activation). Keep the $L'$ coordinates with the largest score. The score depends only on the calibration-set activations, not on the weight matrices.
	\item \emph{Independent contribution (output side).} This is the non-iterative version of the contribution-based criterion used for MLP intermediate-channel pruning in Puzzle~\citep{puzzle}. For a following linear map \(W\) and input activation \(a\), each channel \(i\) is scored independently by the magnitude of its contribution through the following weight column, proportional to \(\mathbb{E}|a_i|\,\|W_{:,i}\|_2\). On the latent output side the following matrix is the shared \(W_{\mathrm{out}}\) and \(a=z^{\mathrm{out}}\), so we keep the \(L'\) latent coordinates with the largest scores.
	\item \emph{Weighted contribution (input side).} The input side has no single following matrix, since each expert applies its own \(U_k\) to the latent vector \(z = W_{\mathrm{in}}x\). We compute the same independent contribution score for each expert using \(U_k\) as the following matrix, aggregate the per-expert scores with calibrated router importance \(\alpha_k\), and keep the top \(L'\) latent coordinates. This is the natural per-coordinate analogue of the Puzzle contribution criterion when the consumer is multi-headed.
	\item \emph{Random and reverse-Minitron controls.} Random selection keeps uniformly sampled latent coordinates. Reverse Minitron keeps the lowest-scoring Minitron coordinates and tests whether the activation score at least identifies coordinates that should not be kept.
\end{itemize}

\paragraph{Projection-based method.} This method instead uses a calibration-weighted low-rank approximation of the composed maps through the latent bottleneck. The \emph{input side} preserves the maps $U_k W_{\mathrm{in}}$ from the model hidden state into each expert, while the \emph{output side} preserves the maps $W_{\mathrm{out}}D_k$ from each expert's intermediate activation back to the model hidden state.

For a calibration set, let $\alpha_k$ be the normalized router importance of expert $k$. Let $h_k=\sigma(U_k W_{\mathrm{in}}x)$ be the intermediate activation entering $D_k$. We define the input activation second-moment matrix $\Sigma_{\mathrm{in}} = \mathbb{E}[xx^\top]$ and the per-expert intermediate activation second-moment matrix $\Sigma_{\mathrm{inter},k} = \mathbb{E}[h_kh_k^\top]$. With an orthonormal input latent projector $P_{\mathrm{in}}\in\mathbb{R}^{L\times L'_{\mathrm{in}}}$, the input-side objective is
\begin{align*}
	\min_{P_{\mathrm{in}}^\top P_{\mathrm{in}} = I}\sum_k \alpha_k\,\mathbb{E}_{x}
	\left\| U_k W_{\mathrm{in}}x
	- U_kP_{\mathrm{in}}P_{\mathrm{in}}^\top W_{\mathrm{in}}x \right\|_2^2
	&=
	\min_{P_{\mathrm{in}}^\top P_{\mathrm{in}} = I}
	\begin{aligned}[t]
		\mathrm{Tr}\hspace{1pt}\!\big(& (I-P_{\mathrm{in}}P_{\mathrm{in}}^\top)\, S_\alpha \\
		& (I-P_{\mathrm{in}}P_{\mathrm{in}}^\top)\, Z\big),
	\end{aligned} \\
	S_\alpha = \sum_k \alpha_k U_k^\top U_k,\quad
	Z = W_{\mathrm{in}} \Sigma_{\mathrm{in}} W_{\mathrm{in}}^\top.
\end{align*}
The analogous output-side objective inserts an orthonormal projector $P_{\mathrm{out}}P_{\mathrm{out}}^\top$ between $W_{\mathrm{out}}$ and each $D_k$:
\begin{align*}
	\min_{P_{\mathrm{out}}^\top P_{\mathrm{out}} = I}\sum_k \alpha_k\,\mathbb{E}_{h_k}
	\left\| W_{\mathrm{out}}D_k h_k
	- W_{\mathrm{out}}P_{\mathrm{out}}P_{\mathrm{out}}^\top D_k h_k \right\|_2^2
	&=
	\min_{P_{\mathrm{out}}^\top P_{\mathrm{out}} = I}
	\begin{aligned}[t]
		\mathrm{Tr}\hspace{1pt}\!\big(& S_W\, (I-P_{\mathrm{out}}P_{\mathrm{out}}^\top) \\
		& Z_\alpha\, (I-P_{\mathrm{out}}P_{\mathrm{out}}^\top)\big),
	\end{aligned} \\
	S_W = W_{\mathrm{out}}^\top W_{\mathrm{out}},\quad
	Z_\alpha = \sum_k \alpha_k D_k \Sigma_{\mathrm{inter},k} D_k^\top.
\end{align*}
On the input side we use the ASVD shortcut~\citep{asvd}: take the top-$L'_{\mathrm{in}}$ left singular vectors of $S_\alpha Z$ as $P_{\mathrm{in}}$ and set $W_{\mathrm{in}}' = P_{\mathrm{in}}^\top W_{\mathrm{in}}$, $U_k' = U_k P_{\mathrm{in}}$. This is the practical approximation used in the implementation; the exact first-order condition for the stated objective is a symmetrized eigenproblem. On the output side we factor $W_{\mathrm{out}} = QR$ via QR decomposition~\citep{qrd}, solve the equivalent PCA problem in the QR coordinates, and take the top-$L'_{\mathrm{out}}$ eigenvectors $V_{\mathrm{out}}$ of $R Z_\alpha R^\top$. The compressed weights are then $W_{\mathrm{out}}' = QV_{\mathrm{out}}$ and $D_k' = V_{\mathrm{out}}^\top R\,D_k$. The input and output compression bases are independent, so $L'_{\mathrm{in}} \neq L'_{\mathrm{out}}$ is allowed.

\paragraph{Setup.}
Numbers in Table~\ref{tab:latent-pruning} are zero-shot, measured immediately after pruning, with no distillation or recovery training. We prune every MoE layer to the same $(L'_{\mathrm{in}}, L'_{\mathrm{out}})$ (homogeneously) and evaluate on a held-out validation split. The teacher here is an earlier Super checkpoint.

\begin{table}[ht]
	\centering
	\small
	\begin{tabular}{@{}lccc@{}}
		\toprule
		Method & LM loss $\downarrow$ & Top-1 acc.\ $\uparrow$ & Routed latent params (\%) \\
		\midrule
		Teacher (no pruning, $L = 1024$) & 0.937 & 0.802 & 100.00 \\
		\midrule
		\multicolumn{4}{l}{\emph{Input-only ($L'_{\mathrm{in}} = 896,\ L'_{\mathrm{out}} = 1024$)}} \\
		Projection-based method & \textbf{0.951} & \textbf{0.799} & \multirow{3}{*}{93.75} \\
		Weighted contribution   & 0.982          & 0.792          & \\
		Minitron                & 0.983          & 0.792          & \\
		\midrule
		\multicolumn{4}{l}{\emph{Output-only ($L'_{\mathrm{in}} = 1024,\ L'_{\mathrm{out}} = 896$)}} \\
		Projection-based method & \textbf{0.957} & \textbf{0.798} & \multirow{3}{*}{93.75} \\
		Independent contribution & 0.961          & 0.796          & \\
		Minitron                & 0.964          & 0.796          & \\
		\midrule
		\multicolumn{4}{l}{\emph{Symmetric ($L'_{\mathrm{in}} = L'_{\mathrm{out}} = 896$)}} \\
		Projection-based method & \textbf{0.979} & \textbf{0.794} & \multirow{5}{*}{87.50} \\
		Random                  & 1.013          & 0.782          & \\
		Weighted contribution   & 1.020          & 0.783          & \\
		Minitron                & 1.024          & 0.782          & \\
		Reverse Minitron        & 1.763          & 0.647          & \\
		\midrule
		\multicolumn{4}{l}{\emph{Symmetric ($L'_{\mathrm{in}} = L'_{\mathrm{out}} = 768$)}} \\
		Projection-based method & \textbf{1.050} & \textbf{0.780} & \multirow{2}{*}{75.00} \\
		Minitron                & 1.234          & 0.738          & \\
		\bottomrule
	\end{tabular}
	\caption{Zero-shot comparison of latent-dimension pruning methods on an earlier Super checkpoint, with no KD or recovery. The projection-based method outperforms the coordinate-selection baselines at each tested latent size. Random and reverse-Minitron are diagnostic coordinate-selection controls. Routed latent params counts only the routed experts' latent up/down projection weights, proportional to $(L'_{\mathrm{in}}+L'_{\mathrm{out}})/(2L)$ for parent latent size $L=1024$; it does not include shared latent projections or unchanged MoE terms.}
	\label{tab:latent-pruning}
\end{table}

\paragraph{Channel selection is not enough; channel mixing is beneficial.} The projection-based method dominates every row in Table~\ref{tab:latent-pruning}: at $896\!\times\!896$ it gives 0.979 LM loss, compared with 1.013--1.024 for the non-reversed coordinate-selection baselines and 1.763 for reverse Minitron. The gap over Minitron widens to 0.18 at $768\!\times\!768$ (1.234 vs.\ 1.050), and on each one-sided block the projection-based method is also strictly best. The reverse-Minitron control confirms that the activation score is not meaningless: the lowest-scoring coordinates are genuinely poor to keep. At the same time, random selection slightly outperforms Minitron at $896\!\times\!896$, suggesting that after removing clearly bad coordinates, the remaining loss-relevant signal is diffuse rather than concentrated on the largest-activation axes. The projection-based method is more general: $P_{\mathrm{in}}$ and $P_{\mathrm{out}}$ are dense, so each retained latent dimension is a linear combination of all original coordinates. For example, on the input side $W_{\mathrm{in}}' = P_{\mathrm{in}}^\top W_{\mathrm{in}}$ and $U_k' = U_k P_{\mathrm{in}}$ produce $L'_{\mathrm{in}}$ new mixed channels rather than $L'_{\mathrm{in}}$ surviving original channels. This ability to rotate the basis explains why channel mixing is more effective than selecting coordinates alone.

\paragraph{Why latent pruning is absent from the final architecture.}
In our setup, adding candidates with latent-dimension pruning to the joint Puzzle search did not improve over spending the same budget on intermediate-channel and top-$k$ pruning. We hypothesize that latent pruning could become beneficial at more aggressive compression rates, once the intermediate-channel axis begins to saturate. A second practical constraint is deployment: NVFP4 MoE kernels in current inference engines (\S\ref{sec:quantization}) require the latent dimension to be a multiple of 512, leaving $L = 1024 \to 512$ as the only deployable latent step. That single step was too aggressive to absorb without per-layer recovery cost beyond what our Puzzle search allowed.

\section{FP8 Results}
Table~\ref{tab:super_turbo_fp8_results} presents the main benchmark accuracies for Nemotron-3-Puzzle-75B-A9B in FP8 precision.

\begin{table}[t]
\centering
\caption{Results of Nemotron-3-Puzzle-75B-A9B in FP8 precision.}
\begin{tabular*}{0.62\linewidth}{@{\extracolsep{\fill}}l|c@{\hspace{1.5em}}}
\toprule
Benchmark & \shortstack{Puzzle-75B-A9B\\FP8} \\
\midrule
\rowcolor{gray!12}\multicolumn{2}{@{}l@{}}{\textbf{General Knowledge}} \\
MMLU-Pro & 82.0 \\
\midrule
\rowcolor{gray!12}\multicolumn{2}{@{}l@{}}{\textbf{Reasoning}} \\
AIME25 (no tools) & 89.4 \\
HMMT Feb25 (no tools) & 92.7 \\
HMMT Feb25 (with tools) & 93.6 \\
GPQA (no tools) & 77.8 \\
GPQA (with tools) & 80.6 \\
LiveCodeBench (v5 2024-07$\leftrightarrow$2024-12) & 80.5 \\
SciCode (subtask) & 39.6 \\
HLE (no tools) & 16.0 \\
\midrule
\rowcolor{gray!12}\multicolumn{2}{@{}l@{}}{\textbf{Agentic}} \\
Terminal Bench (hard subset) & 22.9 \\
\textbf{TauBench V2} &  \\
\quad Airline & 54.5 \\
\quad Retail & 63.4 \\
\quad Telecom & 61.3 \\
\quad Average & 59.7 \\
\midrule
\rowcolor{gray!12}\multicolumn{2}{@{}l@{}}{\textbf{Chat \& Instruction Following}} \\
IFBench (prompt) & 71.9 \\
Scale AI Multi-Challenge & 55.4 \\
Arena-Hard-V2 & 69.8 \\
\midrule
\rowcolor{gray!12}\multicolumn{2}{@{}l@{}}{\textbf{Long Context}} \\
AA-LCR & 56.6 \\
RULER 256k & 95.3 \\
RULER 512k & 94.5 \\
RULER 1M & 92.4 \\
\midrule
\rowcolor{gray!12}\multicolumn{2}{@{}l@{}}{\textbf{Multilingual}} \\
MMLU-ProX (avg over langs) & 77.1 \\
WMT24++ (en$\to$xx) & 85.2 \\
\bottomrule
\end{tabular*}
\label{tab:super_turbo_fp8_results}
\end{table}

\end{document}